\documentclass[iicol, sn-mathphys-num]{sn-jnl}

\usepackage{amsthm}%
\usepackage{mathrsfs}%
\usepackage[title]{appendix}%
\usepackage{manyfoot}%
\usepackage{algorithmicx}%
\usepackage{enumerate}
\usepackage[noend]{algpseudocode}
\usepackage{kantlipsum}
\usepackage{multirow}
\usepackage{amsmath,amssymb,amsfonts}
\usepackage{algorithm}
\usepackage{graphicx}
\usepackage{textcomp}
\usepackage{subcaption}
\usepackage{booktabs}
\usepackage{adjustbox}
\usepackage{xcolor}
\usepackage{siunitx}
\newcommand{\textupdate}[1]{\textcolor{black}{#1}}
\newcommand{\textupdateref}[1]{\textcolor{black}{#1}}
\usepackage{makecell}
\usepackage{multicol,siunitx,enumitem,ragged2e}
\newlength\mylen
\usepackage{orcidlink}
\newcommand{\cmsorcid}[1]{\orcidlink{#1}}
\newcommand{\cmsAuthorMark}[1]{$^{#1}$}
\newcommand\upvDash{\protect\mathpalette{\protect\independenT}{\perp}}
\def\independenT#1#2{\mathrel{\rlap{$#1#2$}\mkern2mu{#1#2}}}
\usepackage{relsize}

\theoremstyle{thmstyleone}%
\theoremstyle{thmstyletwo}%

\theoremstyle{thmstylethree}%
\newtheorem{definition}{Definition}%

\begin{document}

\title[Article Title]{Scalable Temporal Anomaly Causality Discovery in Large Systems: Achieving Computational Efficiency with Binary Anomaly Flag Data}


\author*[1]{\fnm{Mulugeta~Weldezgina} \sur{Asres}}\email{mulugetawa@uia.no}
\author[2]{\fnm{Christian~Walter} \sur{Omlin}}\email{christian.omlin@uia.no}
\author[3]{\fnm{The~CMS-HCAL~Collaboration} \sur{}}\email{}
\equalcont{Full author list is provided at the end of the manuscript.}
\affil[1,2]{\orgdiv{Department of ICT}, \orgname{University~of~Agder}, \city{Grimstad}, \country{Norway}}
\affil[3]{\orgdiv{The CMS Experiment}, \orgname{CERN}, \city{Geneva}, \country{Switzerland}}

\abstract{Extracting anomaly causality facilitates diagnostics once monitoring systems detect system faults. 
Identifying anomaly causes in large systems involves investigating a broader set of monitoring variables across multiple subsystems. 
However, learning graphical causal models (GCMs) comes with a significant computational burden that restrains the applicability of most existing methods in real-time and large-scale deployments.
In addition, modern monitoring applications for large systems often generate large amounts of binary alarm flags, and the distinct characteristics of binary anomaly data---the meaning of state transition and data sparsity---challenge existing causality learning mechanisms.
This study proposes an anomaly causal discovery approach (\textsc{AnomalyCD}), addressing the accuracy and computational challenges of generating GCMs from temporal binary flag datasets. The \textsc{AnomalyCD} presents several strategies, such as anomaly data-aware causality testing, sparse data and prior link compression, and edge pruning adjustment approaches. 
We validate the performance of the approach on two datasets: monitoring sensor data from the readout-box system of the Compact Muon Solenoid experiment at CERN, and a public dataset from an information technology monitoring system.
The results on temporal GCMs demonstrate a considerable reduction of computation overhead and a moderate enhancement of accuracy on the binary anomaly datasets. 
Source code: \href{https://github.com/muleina/AnomalyCD}{https://github.com/muleina/AnomalyCD}
}


\maketitle

\vspace*{-\baselineskip}
\noindent \textbf{Acronyms} 
\begin{description}[font=\normalfont\itshape,labelwidth=\mylen,leftmargin=6.6em,rightmargin=0.0em,noitemsep,nosep]
\item[AD]                              Anomaly Detection                     
\item[ANAC]                            Our Anomaly Aware CI Test             
\item[AnomalyCD]              Our Anomaly CD Approach        
\item[APRC]                            Area under Precision-Recall Curve 
\item[BIC]                             Bayesian information criterion 
\item[Bdeu]                            Bayesian Dirichlet Equivalent Uniform Prior
\item[BN]                              Bayesian Network      
\item[CD]                              Causal Discovery                      
\item[CI]                              Conditional Independence              
\item[CMS]                             Compact Muon Solenoid                 
\item[CP]                              Conditional Probability               
\item[CPD]                             CP Distributions 
\item[DAG]                             Directed Acyclic Graph                
\item[DL]                              Deep Learning                         
\item[FCI]                             Fast Causal Inference                 
\item[FCM]                             Functional Causal Model               
\item[FFT]                             Fast Fourier Transform                
\item[GAE]                             Graph Autoencoder                     
\item[GCM]                             Graphical Causal Model                    
\item[GES]                             Greedy Equivalent Search              
\item[GFCI]                            Greedy Fast Causal Inference          
\item[GS]                              Grow-Shrink                           
\item[HCAL]                            Hadron Calorimeter                    
\item[HE]                              HCAL Endcap Detector                  
\item[HEM]                             HE Minus Hemisphere RBX               
\item[HEP]                             HE Plus Hemisphere RBX                
\item[IAMP]                            Incremental Association               
\item[LHC]                             Large Hadron Collider                 
\item[MMHC]                            Max-Min Hill Climbing                 
\item[MMPC]                            Max-Min Parents and Children        
\item[OD]                              Outlier Detection
\item[PC]                              Peter-Clark   
\item[PCMCI]                           Peter-Clark Momentary CI              
\item[PDAG]                            Partial Directed Acyclic Graph        
\item[QIE]                             Charge Integrating and Encoding       
\item[RBX]                             Readout Box                           
\item[RCA]                             Root-Cause Analysis                   
\item[RM]                              Readout Module                      
\item[SDH]                             Sparse Data Handling
\item[SDLH]                             Sparse Data and Link Handling
\item[SHD]                             Structural Hamming Distance           
\item[SHDU]                            Undirected SHD                        
\item[SiPM]                            Silicon Photomultipliers   
\item[SR]                              Spectral Residual 
\item[TPC]                             Time-Aware PC                         
\item[TS]                              Time Series                           
\item[VTTx]                            Versatile Twin Transmitter           
\item[$x_i, X$]                        Time series Data Variable(s)
\item[$\mathcal{D}$] Dataset, $X \in \mathcal{D}$
\item[$\hat{\mathcal{D}}$]  Dataset after $\mathcal{S}_d$ 
\item[$x_\iota$]    Trend component signal of $x$
\item[$x_\zeta$]    Seasonal component signal of $x$
\item[$x_\epsilon$]   Residual component signal of $x$
\item[$p$]   Period of $x$
\item[$\tau_\text{max}$]  Maximum Time-Lag for CD
\item[$\l_m$]  Length Hyperparameter for $\mathcal{S}_d$ 
\item[$\mathcal{G}$] Graph with Nodes $\mathcal{V}$ and Edges $\mathcal{E}$
\item[$\mathcal{V}$] Set of Nodes of $\mathcal{G}$ 
\item[$\mathcal{E}$] Set of Edges of $\mathcal{G}$ 
\item[$\varepsilon$] Graph Edge Link $\varepsilon \in \mathcal{E}$
\item[$\upsilon, \nu$] Graph Node $\upsilon, \nu \in \mathcal{V}$ 
\item[$\Theta$]    Moving Standard Deviation OD
\item[$\mathcal{L}$]  Trend Deviation OD
\item[$\Phi$]     Spectral OD
\item[$\lambda$]    OD Flag Generation Threshold
\item[$\mathcal{P}$] Probability Function
\item[$\mathbf{P A}_i$] Set of Parent Nodes of $x_i$
\item[$\mathcal{H}$] GCM Extraction Function
\item[$\hat{\mathcal{H}}$] $\mathcal{H}$ using \textsc{AnomalyCD}
\item[$\mathcal{I}$]  CI Test Function
\item[$\hat{\mathcal{I}}$] Anomaly-Aware CI Test Function
\item[$\rho$] Pearson's Correlation
\item[$\mathbf{G}_0$] Initial $\mathbf{G}$, Prior Link Assumption
\item[$\hat{\mathbf{G}_0}$] $\mathbf{G}_0$ after $\mathcal{S}_e$ 
\item[$\Gamma_e$]  Edge Pruning Adjustment Function          
\item[$\Lambda$]   Binary Anomaly Flag Data
\item[$\mathcal{S}_d$] Sparse Data Compression Function    
\item[$\mathcal{S}_e$] Sparse Link Compression Function    
\item[$\mathcal{B}$]  BN Model
\item[$\wp$]    BN Inference Engine

\end{description}

\section{Introduction}
\label{sec:introduction}
Anomaly detection (AD) approaches are commonly employed in industrial monitoring systems to capture anomalies that require attention~\cite{pol2019anomaly, wielgosz2018model, abadjiev2023autoencoder, asres2021unsupervised, asres2022long, mulugeta2022dqm, asres2024dqmtl}. Thus, improving efficiency, safety, and reliability, which will ultimately contribute to reducing maintenance cost and improving quality~\cite{pol2019anomaly, wielgosz2018model, abadjiev2023autoencoder, asres2021unsupervised, asres2022long, mulugeta2022dqm, asres2024dqmtl, rashidi2018data, leonhardt2022pen, ariza2021cms}. 
Discovering anomaly causality from a broader range of sensors, including variables not monitored by pre-trained AD models, is essential for facilitating fault diagnostics through root cause discovery and analysis. 
Causal knowledge of direct and indirect effects, interaction pathways, and time-lags can assist in comprehending the fault nature and modeling physical systems to predict the impact of anomaly occurrence or interventions~\cite{gerhardus2020high}.

The propagating nature of malfunctions makes fault diagnosis challenging in most multivariate processes~\cite{rashidi2018data, asres2024lightweight}. 
Causal knowledge of faults is traditionally acquired through inductive and deductive risk methods that employ variants of failure mode and effect analysis, and fault tree analysis, respectively~\cite{steenwinckel2021flags}. These approaches provide rules for modeling expert knowledge for prior known malfunctions~\cite{martinez2022root}. To effectively utilize these methods, an extensive domain knowledge from many experts is required, which is a time-consuming process in addition to the possible ambiguity and incompleteness in large systems~\cite{steenwinckel2021flags}. 
Data-driven approaches learn causality directly from the data collected by sensor monitoring systems~\cite{glymour2019review, guo2020survey, rashidi2018data, liu2021root, bohmer2020mining, tian2020causal, nauta2019causal, zhou2022root}.  
However, AD and causal discovery (CD) often remain highly intractable due to widely diverse operational modes, disparate data types, and complex fault propagation mechanisms~\cite{liu2021root, guo2020survey, leonhardt2022pen}. 
Recent AD advances for large complex systems focus on specific subsystems due to the curse of dimensionality, data annotation challenges, and the need for accuracy improvement. This results in multiple AD models for monitoring different subsystems \cite{asres2021unsupervised, asres2022long, mulugeta2022dqm, asres2024dqmtl, pol2019anomaly, wielgosz2018model, abadjiev2023autoencoder, ariza2021cms}. 
Identifying the causes of an anomaly requires investigating an extensive set of monitoring variables across subsystems that the trained AD models do not cover. 
An end-to-end framework for anomaly CD that addresses the challenges pertaining to large systems, such as status data from multiple AD systems, dynamics set of monitored variables, and large temporal data, is thus of interest in various domains. 

Graphical causal models (GCMs) provide an intuitive CD reasoning presentation for performing causal inference tasks, such as quantifying the causal strength of variables and root cause analysis (RCA)~\cite{wang2018cloudranger, budhathoki2021did, budhathoki2022causal, assaad2023root}.
Refs.~\cite{runge2018causal, runge2020discovering} present a \textsc{PCMCI} method that is a state-of-the-art for GCM on TS data. The \textsc{PCMCI} combines a constraint-based Peter-Clark (PC) estimator with momentary CI algorithms to achieve good accuracy on large data sets~\cite{runge2018causal, runge2019detecting, runge2020discovering, saggioro2020reconstructing, niu2024comprehensive}.
Although the algorithm incorporates several multi-stage algorithms to reduce computation and improve convergence, the computation cost grows quickly for large datasets, limiting their adoption in real-time applications~\cite{kalisch2007estimating, li2009controlling}. 

We propose an anomaly causal discovery approach (\textsc{AnomalyCD}) that integrates several methods to address computational and accuracy challenges related to generating causal knowledge discovery on multivariate TS anomaly data. 
We employ a temporal online AD method, sparse data handling, causal graph structure learning, and causality inference using probabilistic causal modeling.
Our study focuses on discovering anomaly GCM from time series (TS) binary flags, as most monitoring systems log their data in this format. Binary flag data sets are widely available, and using them for CD lessens the impact of data heterogeneity streamed from diverse AD models, providing data normalization. 
Although there are some recent efforts for inferring GCM from binary data~\cite{amornbunchornvej2023framework} or outlier signals~\cite{budhathoki2022causal}, there is still a gap in addressing the unique challenges of binary anomaly data for large systems. Temporal binary anomaly data has distinctive characteristics different from ordinary categorical or discrete data: 1) flag transitions between normal $0$ and alert $1$ have a meaning of anomaly occurrence and disappearance, and 2) data can exhibit severe sparsity because of long uniform status regimes leading to multicollinearity. 
The existing causality learning mechanisms have yet to be established effectively to handle these characteristics. Their accuracy suffers, and the computation cost grows quickly for large datasets, limiting their adoption in real-time applications. 
The \textsc{AnomalyCD} leverages the accuracy and alleviates the computational burden of the \textsc{PCMCI} algorithm by customizing the conditional independence (CI) test, and sparse data and link compression algorithms to effectively capture temporal anomaly data attributes during graph structure generation. 
We propose simple but novel compression algorithms for binary flag data that substantially reduce the computational cost of the GCM learning process. Moreover, we enable a Bayesian network (BN)~\cite{neapolitan2004learning}, as a probabilistic causal model, to query causality inference on the TS GCM.

We have applied the proposed approaches to multivariate sensor data from the readout box systems of the Hadron Calorimeter (HCAL) of the Compact Muon Solenoid (CMS) experiment at CERN's Large Hadron Collider (LHC). The results establish that our approach accurately detects outlier behaviors and generates causal networks consistent with the actual physical circuit connections and environmental associations. 
We have also assessed and compared the efficacy of our approach with benchmarking state-of-the-art CD methods on a publicly available infrastructure monitoring dataset. 
The \textsc{AnomalyCD} has improved the GCM accuracy and considerably improved the computational efficiency.
Moreover, we have conducted an ablation study on \textsc{AnomalyCD} to demonstrate the contribution of the proposed techniques. 
The key contributions of our study are highlighted below: 
\begin{itemize}
    \item We present an elaborate study on anomaly CD from binary TS data, discussing the impact of data characteristics on GCM and revealing potential limitations of existing approaches in the domain. 
    \item We present \textsc{AnomalyCD} for CD on anomaly flag data that addresses several graphical causal modeling challenges, such as anomaly-aware CI, data and initial non-coinciding edge compression, edge pruning, and adoption of BN for TS causality inference.  
    \item We propose novel compression algorithms that significantly reduce the data size and graph edge search space of sparse binary data. The algorithms achieve promising computational efficiency that leverages speed and accuracy for graph structure learning, paving the way for online CD application. 
\end{itemize}

We discuss the background on CD in Section~\ref{sec:background}. We briefly describe the HCAL readout box system and the public monitoring sensor data sets employed in the study in Section~\ref{sec:datasetdescription}. We present our approaches in Section~\ref{sec:methodology}, and provide results and discussion in Section~\ref{sec:resultsanddiscussion}. We summarize the contribution of our study in Section~\ref{sec:conclusion}. Finally, Appendix \ref{sec:appendix_alg} presents algorithm pseudo-codes and Appendix \ref{sec:rca__causal_results} discusses further results. 
\section{Background}
\label{sec:background}

In this section, we will review anomaly CD approaches and graphical causal modeling, and highlight concepts of BN modeling.

In the realm of system monitoring for complex systems, it is imperative to delve beyond predictive or descriptive machine learning tasks to fully comprehend the cause-and-effect relationship among different variables and systems~\cite{guo2020survey, liu2021root}. 
The investigation of causality is a prominent area of interest in diverse fields, including but not limited to IT systems~\cite{wang2018cloudranger, meng2020localizing}, transportation~\cite{steenwinckel2021flags}, medical science~\cite{spirtes2000causation, chevalley2022causalbench}, meteorology~\cite{guo2020survey}, and social science~\cite{amornbunchornvej2023framework}.
Causal presentation of data is an essential component in RCA that deals with the identification of the underlying root causes and provides an explanation of how the faults are impacting the monitored system~\cite{meng2020localizing, budhathoki2022causal, assaad2023root}. 
Causal models can answer diagnostic questions, such as what would happen if faults occur in particular variables, and predict how specific variables trigger faults in other variables or systems.

The relationship among different variables or components often involves multiple time lags in modern cyber-physical industrial systems. These time lags delay the fault propagation on the causally connected process variables~\cite{chen2022multi}. 
Although temporal data provides valuable context that enhances AD and CD, it can pose specific challenges, such as slower data acquisition rates than the underlying rate of changes in the causal processes, data non-stationarity, and causality heterogeneity and non-linearity (concept drift on the causal relationship)~\cite{glymour2019review, guo2020survey}.

\subsection{Anomaly Causal Discovery}

Recent data-driven approaches for CD and RCA include statistical~\cite{tian2020causal, chen2021detection, rashidi2018data, qin2022root, liu2022fault, de2020generalized}, information theory~\cite{rashidi2018data}, and machine learning algorithms~\cite{chen2021detection, liu2021root, leonhardt2022pen, zhou2022root, chen2022multi}. 
Rashidi~et~al. \cite{rashidi2018data} present kernel principal component analysis (\textsc{Kernel-PCA}) transformation and symbolic transfer entropy to reduce causality analysis computation for RCA fault diagnosis on nonlinear variables. 
The proposed method identifies only potential root causes and does not provide causal interactions among TS variables. 
Tian~et~al.~\cite{tian2020causal} propose convergent cross-mapping based on deterministic system theory to build a causal network for root cause tracing in industrial alarm data. 
Chen~et~al.~\cite{chen2021detection} combine nonlinear chirp mode decomposition with Granger causality to analyze root causes for multiple plant-wide oscillations in a process control system. 
Liu~et~al.~\cite{liu2021root} employ a spatio-temporal pattern network for RCA on TS anomalies in distributed complex systems. The anomalies are detected from changes in the causality dependency networks generated from a model trained on a symbolic representation of the healthy TS data.  
Steenwinckel~et~al.~\cite{steenwinckel2021flags} discuss fusing expert knowledge with a data-driven semantic rule mining (pattern matching using matrix profiling) for AD and RCA applied to predictive maintenance of trains. 
Liu~et~al.~\cite{liu2022fault} discuss fine-tuning Spearman’s rank correlation analysis with domain knowledge rules to build a BN for fault detection and diagnosis. 
Qin~et~al.~\cite{qin2022root} utilize PCA for fault detection and variable importance selection using gradient boosting for RCA. 

Recent approaches have also integrated causality inference into the AD deep learning models~\cite{zhou2022root, chen2022multi}. 
Zhou~et~al.~\cite{zhou2022root} present a temporal convolution and multi-head self-attention networks for CD and a contrastive causal graph for RCA.
Chen~et~al.~\cite{chen2022multi} propose a sparse causal residual neural network model to extract multi-time-lag causality for industrial process fault diagnosis.
Leonhardt~et~al.~\cite{leonhardt2022pen} employ deep graph convolution neural networks to solve the sparse and nonlinear problem of the state space models for RCA. 

GCMs are popular for their intuitive presentation of causality using graphs~\cite{wang2018cloudranger, budhathoki2021did, budhathoki2022causal, assaad2023root}. 
The graphs illustrate influence and effect paths, provide strength and direction of linkage, and enable estimation of influence propagation across their paths.  
These attributes make GCMs convenient for CD and RCA applications~\cite{assaad2023root, budhathoki2022causal}.
Wang~et~al.~\cite{wang2018cloudranger} present \textsc{CloudRanger} that utilizes the non-temporal \textsc{PC} algorithm for TS data to discover the causal graph between anomalies and identify root causes through a random walk on a correlation transition matrix. 
\textsc{MicroCause} in Ref.~\cite{meng2020localizing} enhances the \textsc{CloudRanger} by inferring the GCM using the PCMCI algorithm~\cite{runge2019detecting}. 
\textsc{WhyMDC} in Ref.~\cite{budhathoki2021did} identifies root causes from non-temporal data by searching for changes in causality on a given GCM. 
\textsc{CausalRCA} in Ref.~\cite{budhathoki2022causal} employs the \textsc{Shapley} value~\cite{lundberg2017unified} to quantify contributions and rank root causes of a point anomaly score using non-temporal structural causal models.
\textsc{EasyRCA} in Ref.~\cite{assaad2023root} identifies the root causes of collective anomalies using time of occurrence and divergence graph compared to the GCM built from normal operation.
The study assumes that the graph already exists, which is a challenging assumption to hold in several real-world complex systems.
The above recent approaches for anomaly RCA are not adequately optimized for binary anomaly-flag data CD, as they are limited to non-temporal modeling or require the availability of the GCM network~\cite{wang2018cloudranger, budhathoki2021did, budhathoki2022causal, assaad2023root}. Large complex systems often exhibit challenges, such as temporal fault propagation, the actual fault causal network is not readily available as the number of variables grows, and the fault network can differ from normal operation causal networks.

\subsection{Review on Graphical Causal Discovery}
\label{sec:rca__background_gcm}
The first step of graph-based causality analysis is to construct the GCM of the variables (see Definition~\ref{def:gcm}). 
 \begin{definition}
    \label{def:gcm}
    \textit{Graphical causal model (GCM)} states that if two variables have an edge in between $X \rightarrow Y$ in the directed graph, then $X$ is a direct cause of $Y$; there exist interventions on $X$ that will directly change $Y$ (distribution or value). 
    The edge of GCM between variables $X$ and $Y$ can model 1) a direct causal relationship ($X \rightarrow Y$ or $Y \rightarrow X$), 2) a causal relationship in either direction ($X$\textemdash$Y$), and 3) a non-causal ($X \leftrightarrow Y$) due to external common causes.
\end{definition}

The GCM captures direct dependencies and shared drivers among multiple data variables, the graph nodes.
Data-driven causal graph structure learning methods can broadly be categorized into: 
\begin{itemize}
    \item \textit{Constraint-based}: Relies on conditional independence (CI) relationships $X \upvDash Y | Z$ ($X$ independent of $Y$ condition on $Z$) to infer the causal directed acyclic graph (DAG) structure. Some of the popular methods include PC~\cite{spirtes2000causation}, grow-shrink (GS)~\cite{margaritis2003learning}, incremental association (IAMP)~\cite{tsamardinos2003algorithms}, max-min parents and children (MMPC)~\cite{tsamardinos2003time}, and fast causal inference (FCI)~\cite{spirtes2000causation}). 
    The PC algorithm in Refs.~\cite{spirtes2000causation, colombo2014order} is a widely used method that builds a causal graph by adding edges based on CI tests. It learns a partial DAG (PDAG) representing the dependencies based on the causal Markov condition and the faithfulness assumption. 
    With no latent confounder, two variables are directly causally related with an edge in between if and only if there is no subset of the remaining variables conditioning on which they are independent.
   
    \item \textit{Score-based}: Employs optimization search for causal DAG structure to the observed data based on a scoring metric~\cite{tsamardinos2006max, glymour2019review, guo2020survey, gerhardus2020high}; e.g., hill-climbing search (HC)~\cite{koller2009probabilistic}, and greedy equivalent search (GES)~\cite{chickering2002optimal, ramsey2015scaling}. 
    These methods explore the space of PDAG structures and minimize a global score---e.g., the \textit{Bayesian information criterion (BIC)} and \textit{Bayesian Dirichlet equivalent uniform prior (BDeu)}---using add, remove, and reverse edge operators to return the optimal structure~\cite{chickering2002optimal, ramsey2015scaling}. 
    
    \item \textit{Hybrid}: Combines constraint- and score-based algorithms to enhance accuracy and computational efficiency; e.g., max-min hill climbing (MMHC)~\cite{tsamardinos2006max}, and greedy FCI (GFCI)~\cite{ogarrio2016hybrid}. The approaches build a first sketch of the graph using CI tests, and then orient the edges through score metrics or vice versa.
   
    \item \textit{Functional causal model (FCM)}: is a recent CD approach that represents the effect as a function of the causes and independent noise terms~\cite{tsamardinos2006max, ogarrio2016hybrid}; e.g., causal additive models~\cite{buhlmann2014cam} and causal generative neural networks~\cite{goudet2018learning} and others~\cite{glymour2019review, guo2020survey}. 
    FCM captures the asymmetry between cause and effect variables, representing the effect $Y$ as a function of the direct causes $X$ and noise factor $\vartheta$ as: 
    \begin{equation} 
        Y = f_ \theta(X, \vartheta; \theta),
    \end{equation}
    where $\vartheta$ is the noise term that is assumed to be independent of $X$, the function $f$ explains how $Y$ is generated from $X$, and $\theta$ is the parameter set of $f$.

\end{itemize}

Structured reviews on graph learning methods for GCM are available in Refs.~\cite{glymour2019review, guo2020survey}, and a more detailed explanation of structural causal model properties in a textbook~\cite{peters2017elements}. 
We limit our discussion to studies related to TS data as in Refs.~\cite{liu2020simplified, shojaie2022granger, gerhardus2020high, runge2018causal, runge2019detecting, runge2020discovering, saggioro2020reconstructing, nauta2019causal, zhou2022root, chen2022multi, biswas2022statistical}.

\subsubsection{Causal Discovery in Time Series}

\begin{figure}[!h]
\centering
\includegraphics[width=0.9\linewidth]{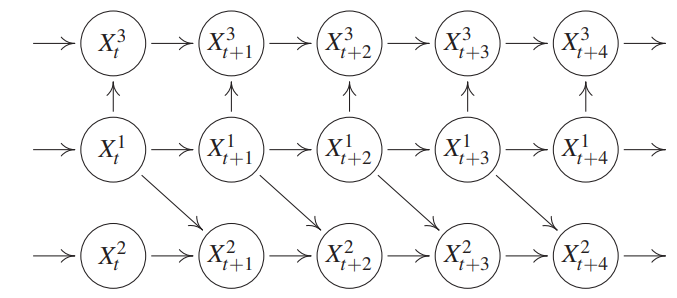}
\caption{A TS with time lag effect $\mathbf{x}^1_{t - 1} \rightarrow \mathbf{x}^2$ and instantaneous effect $\mathbf{x}^1_t \rightarrow \mathbf{x}^3_t$~\cite{peters2017elements}.}
\label{fig:rca__rca_ts_lag_causal_unrolled_plot}
\vspace*{-\baselineskip}
\end{figure}

Graphical causal modeling for TS data is a growing study area in several scientific disciplines~\cite{runge2018causal, liu2020simplified}.
TS data have become abundant in several real-world domains with the proliferation of sensor networks, but finding the causal dynamics in such data is challenging due to the temporal dependency, data non-linearity and non-stationarity, concept drift over time and varying data rates~\cite{glymour2019review}.
Time series GCM approaches need to capture time-lag causality (see Figure~\ref{fig:rca__rca_ts_lag_causal_unrolled_plot}). 

References~\cite{runge2018causal, runge2019detecting, runge2020discovering, saggioro2020reconstructing, gerhardus2020high} introduce and extend variants of PCMCI, an extension of the PC algorithm~\cite{spirtes2000causation} leveraged with false-positive cleaning momentary conditional independence for TS CD. The PCMCI methods with linear and non-linear conditional independence tests outperform state-of-the-art techniques in causality detection on large TS data sets across a range of research fields~\cite{runge2019detecting}.
Gerhardus~et~al.~\cite{gerhardus2020high} propose Latent-PCMCI that relaxes the causal sufﬁciency assumption of PC and extends PCMCI to enhance recall CD with unknown latent confounders using FCI. Saggioro~et~al.~\cite{saggioro2020reconstructing} extend the PCMCI to handle non-stationarity with regime-dependent GCMs using a time-windowing method. 
A slightly different approach is adopted in the time-aware PC (TPC) that employs the PC algorithm for TS by considering time delay, bootstrapping, and pruning~\cite{biswas2022statistical}. The approach unrolls the TS data (adding new nodes with time delay tags) and generates a DAG by applying a set of conditions.
Nauta~et~al.~\cite{nauta2019causal} present a temporal CD framework that consists of independent prediction convolutional networks for each variable, and attention scores are employed to infer the causal links among the variables. 
Training prediction models for each variable can constrain scalability when the number of variables increases, which is often the case in large complex systems. 

\subsection{Bayesian Networks}

Probabilistic parameterization methods, such as BNs, allow flexible and faster querying for causal inference after building the causal topology structure. 
A BN is a probabilistic graphical model representing variables and their conditional dependencies through a DAG~\cite{neapolitan2004learning}. BNs are parameterized using \textit{conditional probability distributions} (CPD), and each node $x_i$ in the network is modeled as $\mathcal{P}(x_i | \mathbf{P A}(x_i))$, where $\mathcal{P}$ is a probability operator, and $\mathbf{P A}(x_i)$ represents the parents of the node $x_i$ in the network.

\begin{definition}
\label{def:bn}
\textit{A Bayesian network (BN)} is a probabilistic representation of joint distributions over the variables using a DAG model. The CPD is computed using the DAG from data using Bayes and the chain rules of probability as:
\begin{equation}
\mathcal{P}(A, B, C) = \mathcal{P}(A| B, C) \mathcal{P}(B|C) \mathcal{P}(C).
\end{equation}
\end{definition}
where the equation shows that the joint distribution of the variables is the sum of the CPDs in the network. 
Methods, such as the \textit{Maximum likelihood estimation} and \textit{Bayesian parameter estimator}, are widely employed BN learning techniques~\cite{koller2009probabilistic}.
The maximum likelihood estimates the CPDs simply using the relative frequencies with which the variable states have occurred. 
The Bayesian parameter estimator mitigates the overfitting issue of relative frequencies (e.g., not fully representative data) using prior histogram counts, such as the \textit{BDeu} to reflect a uniform distribution over possible values~\cite{koller2009probabilistic}.

\section{Dataset Description}
\label{sec:datasetdescription}

This section briefly describes the data sets discussed in our CD study. 

\subsection{Dataset from the HCAL}
\label{sec:dataset_rbx}
\begin{figure}[!htbp]
\centering
\begin{subfigure}[]{\linewidth}
\centering
\includegraphics[width=1\linewidth, scale=1]{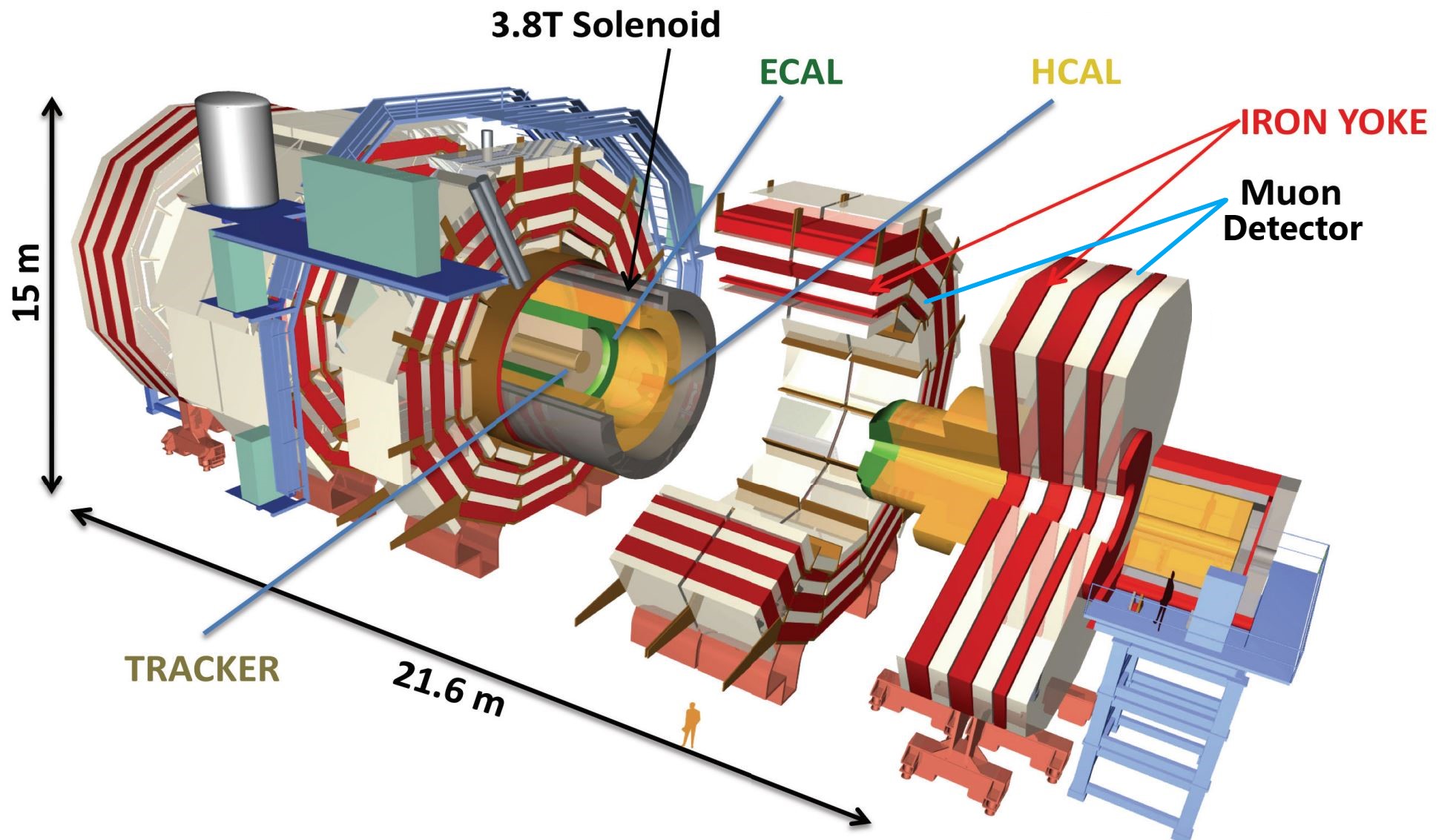}
\caption{}
\end{subfigure}
\caption{Schematic of the CMS experiment~\cite{focardi2012status}
}
\label{fig:cms_diagram_with_hcal}
\vspace*{-\baselineskip}
\end{figure}

\begin{figure}[!htbp]
\centering
\includegraphics[width=1\linewidth]{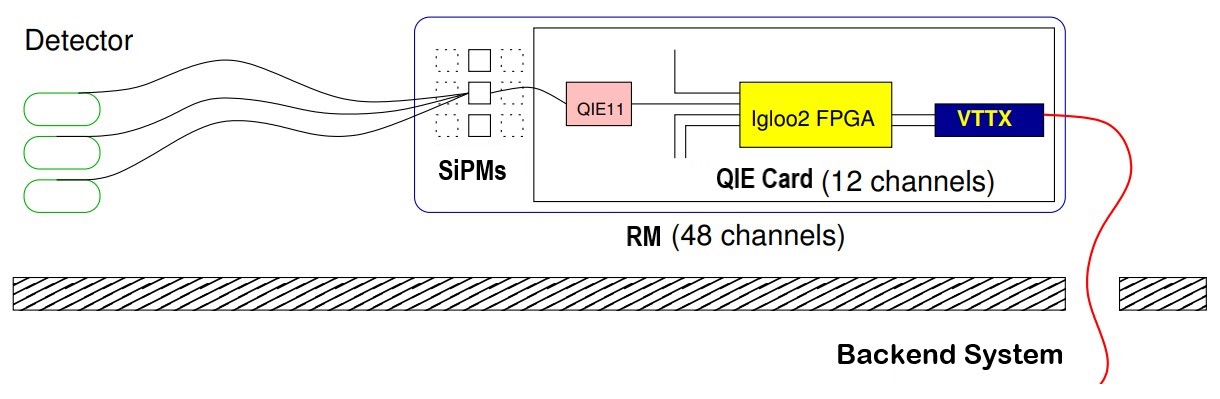}
\caption{The frontend electronics of the HE data acquisition chain, including the SiPMs, the frontend readout cards, and the optical link connected to the back-end electronics~\cite{strobbe2017upgrade}. 
Each readout card contains twelve QIE11 for charge integration, an Igloo2 FPGA for data serialization and encoding, and a VTTx optical transmitter
}
\label{fig:HE_data_acquisition_system_chain}
\vspace*{-\baselineskip}
\end{figure}

The HCAL is a specialized calorimeter that captures hadronic particles during a collision event in the CMS experiment (see Figure~\ref{fig:cms_diagram_with_hcal}) \cite{collaboration2008cms, cms2023development}. 
The primary purpose of the HCAL is to measure the energy of hadrons by absorbing their energy and converting it into measurable signals. 
The calorimeter is composed of brass and plastic scintillators, and the scintillation light produced in the plastic is transmitted through wavelength-shifting fibers to \textit{Silicon photomultipliers}~(SiPMs) (see Figure~\ref{fig:HE_data_acquisition_system_chain}) \cite{strobbe2017upgrade, cms2023development}. 
The HCAL front-end electronics consist of components responsible for sensing and digitizing optical signals of the collision particles. 
The front-end electronics are divided into sectors of \textit{readout boxes} (RBXes) that house and provide voltage, backplane communications, and cooling to the data acquisition electronics.

The HCAL is made of four subsystems (subdetectors): the HCAL Endcap~(HE), the HCAL Barrel~(HB), the HCAL Forward~(HF), and the HCAL Outer~(HO)~\cite{mans2012cms}. 
The front-end electronics of the HE, the main use-case of our study, is made of 36 RBXes arranged on the plus and minus hemispheres of the CMS detector~\cite{mans2012cms, baiatian2008designhe}. 
Each RBX houses four \textit{readout modules}~(RMs) for signal digitization~\cite{strobbe2017upgrade}. Each RM has a SiPM control card, 48~SiPMs, and four readout cards---each includes 12 \textit{charge integrating and encoding chips}~(QIE11) and a \textit{field programmable gate array} (Microsemi Igloo2 FPGA). 
The QIE integrates charge from each SiPM at 40~MHz, and the FPGA serializes and encodes the data from the QIE. The encoded data is optically transmitted to the backend system through the CERN \textit{versatile twin transmitter} (VTTx) at 4.8~Gbps.

The HCAL-RM dataset includes sensor readings from the RMs of the thirty-six RBXes in the HE detector.
Each RM has twelve diagnostic sensors, four from the SiPM control card and eight from four QIE cards (see Table~\ref{tbl:rca_gcm__rm_var_desc}). 
The dataset was obtained from the HCAL online software monitoring system (the ngCCM server) from 01/08/2022 to 30/11/2022. The monitoring sensor data comprises four-month data of 20.7M samples, around 12K per sensor per RM. The dataset contains irregularly sampled and considerably sparse data; a few samples are logged every eight hours for the SiPM control card and every two hours for the QIE card sensors. 
We utilize data from all four RMs (RM-1, RM-2, RM-3, and RM-4) of the RBX-HEP07, interpolated at one-minute interval for the online AD and CD evaluation. The RBX-HEP07 is one of the RBX with diverging behavior as discussed in the multivariate interconnection analysis study in Ref.~\cite{asres2024lightweight}.

\begin{table*}[!htbp] \small
\centering
\caption{HE-RM monitoring sensor data variables description}
\noindent
\resizebox{1.0\linewidth}{!}{
\begin{tabular}{llll}
\toprule
\textbf{No.} & \textbf{Notation} & \textbf{Variable Name} & \textbf{Description} \\
\midrule
1            &      SPV             &     PELTIER\_VOLTAGE\_F                   &  Voltage of the SiPM Peltier temperature controller              \\
2            &     SPC              &      PELTIER\_CURRENT\_F                  &  Current of the SiPM Peltier temperature controller               \\
3            &     SRT              &      RTDTEMPERATURE\_F                  &  SiPM control card temperature averaged over 50 samples              \\
4            &     SCH              &      HUMIDITY\_F                  &  SiPM control card humidity              \\
5            &     Q1H              &      1-B-SHT\_RH\_F                  &  QIE card 1 humidity              \\
6            &     Q2H              &      2-B-SHT\_RH\_F                  &  QIE card 2 humidity              \\
7            &     Q3H              &      3-B-SHT\_RH\_F                  &  QIE card 3 humidity              \\
8            &     Q4H              &      4-B-SHT\_RH\_F                  &  QIE card 4 humidity              \\

9            &     Q1T              &      1-B-SHT\_TEMP\_F                  &  QIE card 1 temperature              \\
10            &     Q2T              &      2-B-SHT\_TEMP\_F                  &  QIE card 2 temperature              \\
11            &     Q3T              &      3-B-SHT\_TEMP\_F                  &  QIE card 3 temperature              \\
12            &     Q4T              &      4-B-SHT\_TEMP\_F                  &  QIE card 4 temperature              \\
\botrule
\end{tabular}
}
\label{tbl:rca_gcm__rm_var_desc}
\vspace*{-\baselineskip}
\end{table*}

\begin{figure}[]
\centering
\includegraphics[width=1\linewidth]{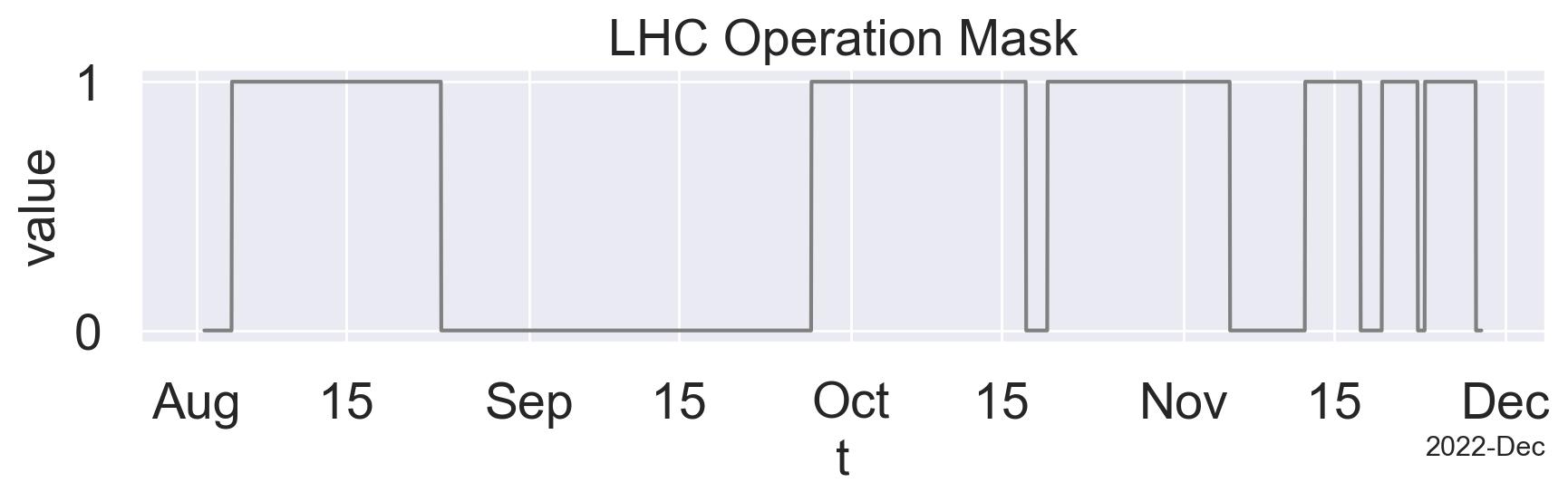}
\caption{The active mask of the LHC operation status from August to December of 2022. The active mask ($1$) refers to the LHC's normal operation run during collision experiment or idle, whereas the inactive mask ($0$) for non-physics operation states, e.g., the LHC's technical stop and maintenance}
\label{fig:rca__lhc_ops_mask_aug_dec_2022}
\vspace*{-0.5\baselineskip}
\end{figure}

\begin{figure*}[htbp]
\centering
\includegraphics[width=1\linewidth]{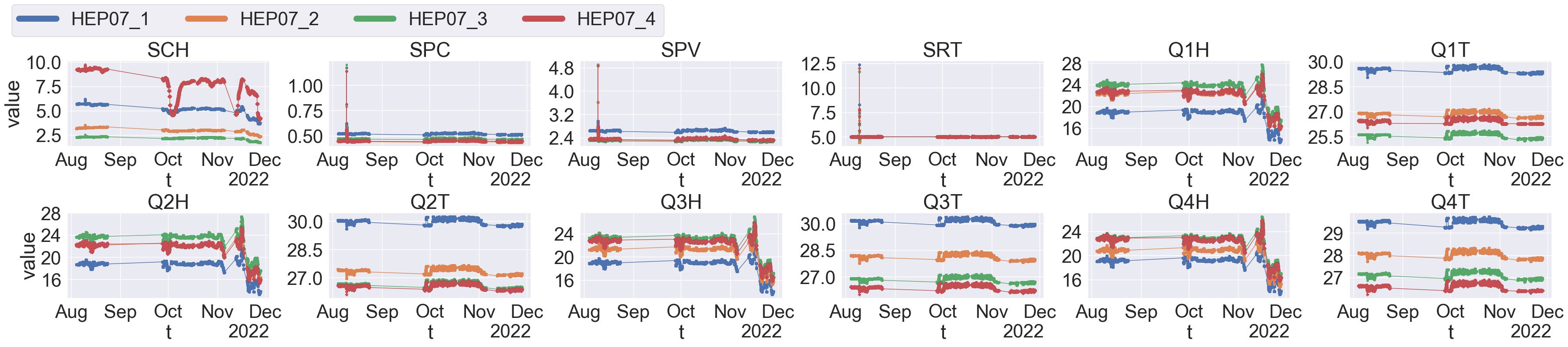}
\caption{Sensor TS data from all RMs of the RBX-HEP07. The HEP07\_i denotes the $i^{\text{th}}$ RM of the RBX}
\label{fig:rca__rm_sensor_data_HEP07_rm_all__01-08__31-11-2022_1min_ts_signal}
\vspace*{-\baselineskip}
\end{figure*}

Several extended reading gaps exist in the data due to various non-physics activities on the LHC. 
We generated a reading mask, shown in Fig.~\ref{fig:rca__lhc_ops_mask_aug_dec_2022}, to filter out the irrelevant regions of the LHC and ensure any detected anomalies are during the normal operation of the collider. 
The data comprise short-lived transient and time-persistent anomalies, such as trend drifts (see Figure~\ref{fig:rca__rm_sensor_data_HEP07_rm_all__01-08__31-11-2022_1min_ts_signal}). We employ data from multiple RMs to capture the global causality of the RM of the HE. 

\subsection{Dataset from EasyVista}
\label{sec:dataset_easyvista}

EasyVista~\cite{easyvista2023} has provided a multivariate sensor dataset from their information technology monitoring system and made the data publicly available in Ref.~\cite{EasyRCA2023git}. 
The dataset consists of eight TS variables collected with a one-minute sampling rate (see Table~\ref{tbl:easyvista__data_var_desc}).   
Following Assaad et al.~\cite{assaad2023root}, we utilize a data segment, indexed from $45,683$ to $50,000$, for our CD evaluation, and each sensor is considered anomalous, exhibiting collective anomalies with close time of appearance and duration. 

\begin{table*}[!htbp] \small
\centering
\caption{EasyVista monitoring data variables description}
\noindent
\resizebox{1\linewidth}{!}{
\begin{tabular}{lll}
\toprule
\textbf{No.} & \textbf{Notation} & \textbf{Description} \\
\midrule
1 & PMDB &  Extraction of message information received by the Storm ingestion system. \\
2 &  MDB &  Activity of message orientation based on the type. \\
3 &  CMB &  Activity of extraction of metrics from messages. \\
4 &  MB &  Activity of insertion of data in a database. \\
5 &  LMB &  Activity of updates of the last metrics values in Cassandra. \\
6 &  RTMB &  Activity of searching to merge data with the check message bolt information. \\
7 &  GSIB &  Activity of insertion of historical status in the database. \\
8 &  ESB &  Activity of writing data in Elasticsearch. \\ 
\botrule
\end{tabular}
}
\label{tbl:easyvista__data_var_desc}
\vspace*{-0.5\baselineskip}
\end{table*}
\section{Methodology}
\label{sec:methodology}

This section presents our proposed \textsc{AnomalyCD} approach for anomaly CD from TS data sets. 

\begin{figure*}[]
\centering
\includegraphics[width=0.8\linewidth]{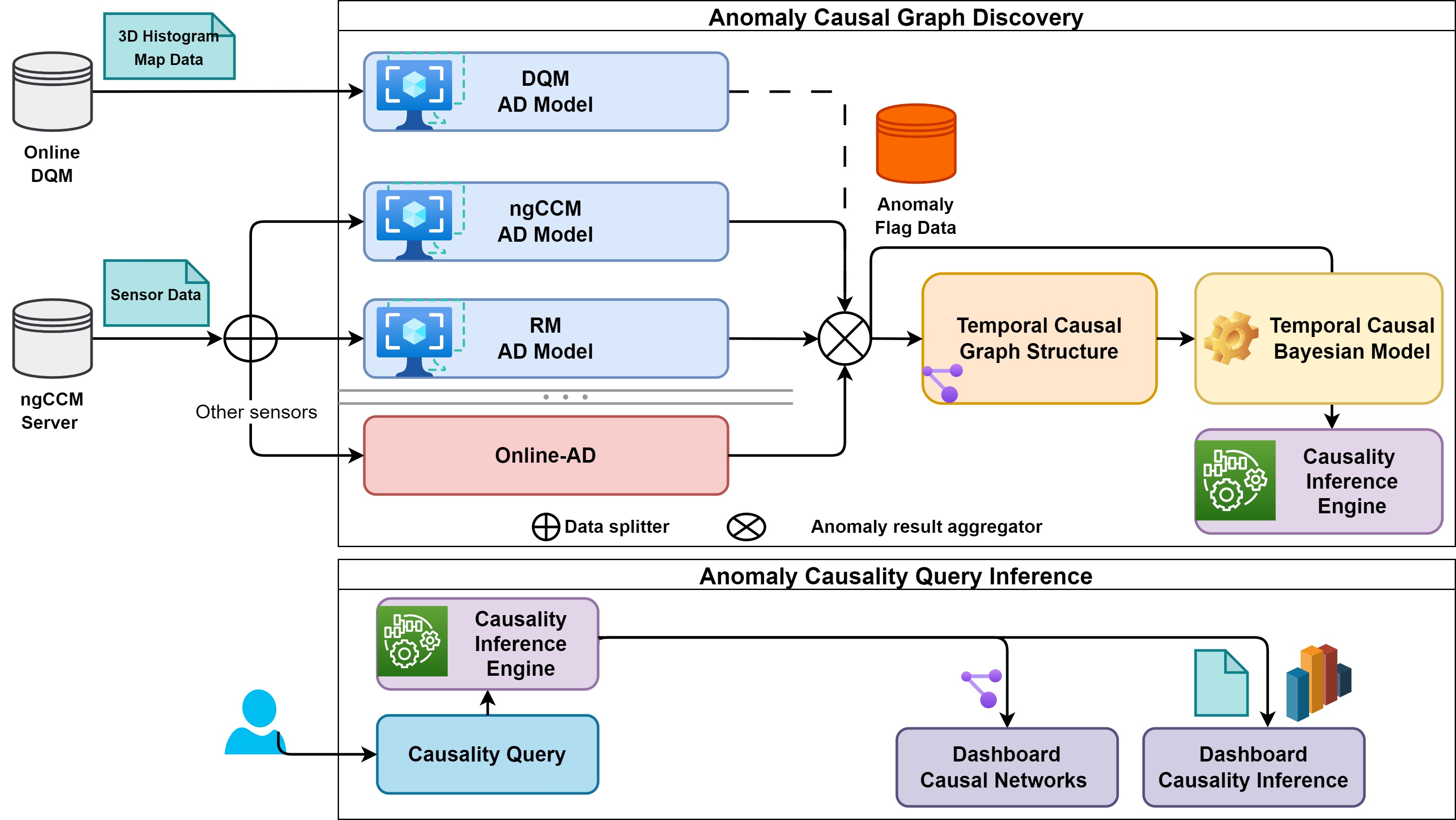}
\caption{\textsc{AnomalyCD}: our anomaly CD approach with use-case of monitoring the HCAL systems. The online-AD detects anomalies on variables that are not actively monitored by the trained AD models of the HCAL, such as the \textit{next-generation clock and control module} (ngCCM), \textit{readout module} (RM), and \textit{data quality monitoring} (DQM) AD models~\cite{asres2021unsupervised,asres2022long, mulugeta2022dqm,asres2024dqmtl,asres2024lightweight}
}
\label{fig:rca__online_rca-rca_online_main}
\end{figure*}

We present the \textsc{AnomalyCD} for causality diagnostics that addresses the challenges of GCM of binary anomaly data sets. The system comprises two main modules, i.e., causal graph discovery and causality inference (see Fig.~\ref{fig:rca__online_rca-rca_online_main}). The causal graph discovery generates temporal causal networks and trains a BN inference model on the anomaly data streamed from AD systems. The causality inference modules handle user queries with observation conditions to provide causality and conditional probabilities using the BN model. 

We have previously trained deep-learning AD models on frontend electronics of the HE detector, including the \textit{next-generation clock and control module} (ngCCM), RM, and \textit{data quality monitoring} (DQM)~\cite{asres2021unsupervised, asres2022long, mulugeta2022dqm,asres2024dqmtl}.  
The HCAL comprises several electronic components, and it is often essential to investigate multiple systems in the pipeline to diagnose system faults. 
We incorporate online temporal AD that detects anomalies on variables that are not actively monitored by trained models in Refs.~\cite{asres2021unsupervised, asres2022long, mulugeta2022dqm, asres2024dqmtl}. Since previous works in Ref.~\cite{asres2021unsupervised, asres2022long, mulugeta2022dqm, asres2024dqmtl} have already discussed the above-trained AD model, this study will focus on the proposed online AD and CD approaches.

\subsection{Online Anomaly Detection}
\label{sec:rca__method_online_od} 

We present an online TS AD algorithm to detect outlier temporal patterns and generate anomaly flags for sensor variables that trained DL models do not actively monitor. 
Different types of variations can occur in TS data, including long-term trends, seasonal changes, periodic fluctuations, and non-random sources of variations~\cite{karim2019multivariate}; these variations can impact the modeling approach and algorithm choices. 
Building a generic one-fits-all approach is challenging as the requirements depend on signal characteristics and target application. 
We propose online AD on univariate TS data to capture typical points and collective anomalies, including transient changes in time and frequency, and gradual signal trend drifts (see Algorithm \ref{alg:rca__online_outlier_detection} in \textupdate{Appendix \ref{sec:appendix_alg}}). 
The approach consists of an ensemble of time- and frequency-domain AD algorithms, presented at \textupdateref{line 2} and \textupdateref{line 3} in Algorithm \ref{alg:rca__online_outlier_detection}, respectively. The final binary AD flag data $\Lambda$ is prepared using bit-wise union from each AD algorithm:
\begin{equation}
        \Lambda = \Lambda_\theta \cup \Lambda_\iota  \cup \Lambda_\phi,
\end{equation}
where $\Lambda_\theta$, $\Lambda_\iota$, and $\Lambda_\phi$ denote the temporal transient, temporal trend, and spectral AD flags, respectively.  

\subsubsection{Time-Domain AD}

The Temporal AD detects transient temporal outliers and degradation in temporal trends on decomposed TS data (see in Algorithm \ref{alg:rca__online_outlier_detection_time_domain}). 
We apply \textit{seasonal and trend decomposition} algorithm from Ref.~\cite{cleveland1990stl} to estimate the trend and residual signals, represented by $x_\iota$ and $x_\epsilon$, respectively:
\begin{equation}
\begin{aligned}
\label{eq:ts_decomp_add}
    x(t) &= x_\iota(t) + x_\zeta (t) + x_\epsilon(t), \\
   x_\iota (t) &= (h_p \ast x)(t),\\
   &\text{where}~h_p (t) = \frac{1}{p} [1,1,...1]. \\
\end{aligned}
\end{equation}
The additive trend $x_\iota$ is obtained by applying a convolution ($\ast$) filter $h_p$ (a moving average with period $p$) to the data. The average of the de-trended series $x(t) - x_\iota(t)$ (after trend removal) over each period returns the seasonal component $x_\zeta$. The final remaining component of the TS becomes the residual error $x_\epsilon$.
We employ \textit{fast Fourier transform} (FFT) on $\Delta x(t): x(t) - x(t-1)$ to enhance the accuracy of the period $p$ estimation in the presence of additive trends (see line \textupdateref{3} in Algorithm \ref{alg:rca__online_outlier_detection_time_domain} in \textupdate{Appendix \ref{sec:appendix_alg}}).
The decomposition method in Eq. \eqref{eq:ts_decomp_add} assumes an additive trend and single seasonality or cyclic pattern. Other techniques, such as multiplicative trend or multi-seasonal component decomposition, can also be utilized depending on the expected normal signal characteristics~\cite{bandara2021mstl}.  

We utilize a sliding z-score algorithm to generate $\Lambda_\theta$ from $x_\epsilon$ for the transient temporal AD (see $\Theta$ at \textupdateref{line 5} in Algorithm \ref{alg:rca__online_outlier_detection_time_domain}). The $x_\epsilon$ signal is normalized by subtracting the average $\mu_w$ and dividing it by the standard deviation $\sigma_w$ using a sliding time-window ($w_ \theta$) to generate the AD score $\lambda_\theta$ (see line \textupdateref{11}). The sliding window localizes the AD to the signal characteristics at adjacent TS data points. 
The $\mu_w$ and $\sigma_w$ can be sensitive to strong outliers in the data, reducing AD efficacy, and it exacerbates for smaller sliding windows. 
Thus, we utilize data quantiles $Q=[10\%, 90\%]$ along with median centering, instead of mean, to reduce the influence of outliers on the $\mu_w$ and $\sigma_w$ estimation (see line \textupdateref{10}). 
Finally, we apply a threshold on the AD score to generate the AD flags $\Lambda_\theta$ (see line \textupdateref{12}). 

We develop a cumulative-based algorithm to detect drifts and generate $\Lambda_\iota$ from the trend signal $x_\iota$ (see $\mathcal{L}$ at \textupdateref{line 6} in Algorithm \ref{alg:rca__online_outlier_detection_time_domain}). 
We estimate first the trend score $\lambda_\iota$ using a cumulative sum on the first-order difference of the $x_\iota$ (see lines \textupdateref{13--21}). The $\mathcal{L}$ is designed to detect gradual signal drifts, and the trend score thus gets reset when during significant signal jumps or drops, often occurs during system configuration or maintenance updates. The z-score and spectral methods capture such sudden shifts.
We apply the threshold $\alpha_\iota$ to get the trend drift AD flags $\Lambda_\iota$ (see line \textupdateref{21}).

\subsubsection{Frequency-Domain AD}
We employ spectral residual (SR) saliency detection to identify frequency spectrum or data rate change outliers (see Algorithm \ref{alg:rca__online_outlier_detection_freq_domain} in \textupdate{Appendix \ref{sec:appendix_alg}}). The SR method has been employed as a preprocessing technique for cleaning outliers and transforming data in semi-supervised AD research, as demonstrated in Refs.~\cite{asres2021unsupervised} and~\cite{ren2019time}, respectively. 
The SR algorithm consists of three major steps for a given univariate data $x(t)$~\cite{ren2019time}: 1) FFT $\mathfrak{F}$ to get the log amplitude spectrum; 2) calculation of spectral residual; and 3) inverse FFT $\mathfrak{F}^{-1}$ that transforms the data back to the time domain and generate saliency AD scores:
\begin{equation}
\begin{aligned}
     A(f),~P(f)&=\mathfrak{F}\left( x(t) \right), \\
    ~\bar{A}(f)&=\log \left( A(f) \right),
    ~\bar{A}_h(f)=h_q(f) * \bar{A}(f), \\
     R(f)&=\bar{A}(f)- \bar{A}_h(f), \\
 \Phi (x(t))&=\left\|\mathfrak{\Im}^{-1}(\exp \left( R(f)+i P(f) \right) )\right\|,
\end{aligned}
\end{equation}
where $A(f)$ is the amplitude spectrum; $P(f)$ is the corresponding phase spectrum; $\bar{A}(f)$ is the log representation of $A(f)$; and $\bar{A}_h(f)$ is the average spectrum of $\bar{A}(f)$ which can be approximated by convoluting with $h_q(f)$, where $h_q(f) = \frac{1}{q} [1,1,...1]$ is averaging filter with an $1 \times q$ vector.
$R(f)$ is the spectral residual, the difference between the log spectrum $\bar{A}(f)$ and the averaged log spectrum $\bar{A}_h(f)$. 
The data is transferred back to the time domain via $\mathfrak{F}^{-1}$ to get the saliency signal $\phi$.
We apply a threshold $\alpha_\phi$ to detect anomaly on $\lambda_\phi$, normalized value of $\phi$ (see line \textupdateref{3}) and generate the flags $\Lambda_\phi$ (see line \textupdateref{4}). 

\subsection{Anomaly Causal Discovery}
\label{sec:rca__method_rcd_alg}

Graphical anomaly CD generates a DAG representing the causal interaction among the monitored variables. 
Computing a DAG for the temporal anomaly data in large systems pose computational and accuracy challenges due to large data size, high multidimensional and binary data sparsity leading to multicollinearity. 
We incorporate several methods to address these challenges using 1) data compression using multivariate sparse handling algorithm, 2) sparsity-driven prior link assumption compression, 3) anomaly flag-aware CI test, and 4) post-processing link adjustment.

The approach infers the temporal causal graph structure among monitoring variables or sensors using time-lagged and contemporaneous CD algorithms. We employ PCMCI algorithm for its accuracy in temporal GCM on large data sets~\cite{runge2020discovering,niu2024comprehensive} and propose augmentation algorithms to enhance its effectiveness for the particular challenges of large binary anomaly data.
The PCMCI can result in spurious links due to errors in estimation when dealing with a long sequence of uniform binary anomaly regions among sensors due to multicollinearity~\cite{assaad2023root,castri2023enhancing, tarraga2024causal}. 
The multicollinearity makes it difficult to assess the relative importance of highly correlated variables because of overlapping information, compromising the ability to infer a true GCM.
Through sparse data and initial link assumption compression, CI testing sensitive to binary flag transitions ($0 \rightarrow 1$), and spurious edge pruning, we alleviate the computational cost and enhance the accuracy of causal graph building.

We leverage the PCMCI~\cite{runge2020discovering, gerhardus2020high} to learn the graph skeleton $\mathcal{G}(\mathbf{V}, \mathbf{E})$ from the $\Lambda$ data, where $\mathbf{V}$ is the set of sensor node vertices $\upsilon, \nu \in \mathbf{V}$, and $\mathbf{E} \subseteq \mathbf{V} \times \mathbf{V}$ is the set of edges $\varepsilon (\upsilon,\nu) \in \mathbf{E}$. 
The PCMCI, a constraint-based GCM methods, rely on CI tests to estimate links among variables.
The independence score function $\mathcal{I} (X, Y)$ answers CI queries of the form $X \upvDash Y | Z$ on a given dataset $\mathcal{D}$, where the variables are assumed to be generated independently from some (unknown) Bayesian system as:
\begin{equation}
\mathcal{I} (X, Y): \mathcal{P}(X, Y | Z) = \mathcal{P}(X | Z)\mathcal{P}(Y | Z),
\end{equation}
where the trade-off of controlling the Type I errors (false positives that reject true independence) and Type II errors (false negatives that accept false independence) is obtained by a significance level threshold $p_v > \alpha_p$ given an independence measuring function $f_\mathcal{I}(\mathcal{D})$, where $p_v=f_\mathcal{I}(\mathcal{D})$, probability of observing independence.
 
Let the PCMCI GCM extracting function is $\mathcal{H} (\mathcal{F}_t, \mathcal{I}, \mathcal{D}, \mathcal{G}_0)$, where $\mathcal{F}_t$ is the temporal CD algorithm, $\mathcal{I}$ is the CI test function, $\mathcal{D} \in \mathcal{B}^{n\times N}$ is the TS dataset with $N$ variables and $n$ data samples. The $\mathcal{G}_0$ is the initial search graph (prior link assumption) with all possible edges. The \textsc{AnomalyCD} GCM extractor $\hat{\mathcal{H}}$ is formulated as:
\begin{equation}
\begin{aligned}
    \hat{\mathcal{H}} (\mathcal{F}_t, \mathcal{I}, \mathcal{D}, \mathcal{G}_0) = \Gamma_e\left( \mathcal{H} (\mathcal{F}_t, \hat{\mathcal{I}}, \hat{\mathcal{D}}, \hat{\mathcal{G}_0}) \right),
\end{aligned}
\end{equation}
where $\hat{\mathcal{I}}$ is anomaly-aware CI testing function, $\hat{\mathcal{D}}=\mathcal{S}_d(\mathcal{D}, l_m)$ is compressed data after sparse data handling using $\mathcal{S}_d$ and compression length hyperparameter $l_m$, and $\hat{\mathcal{G}_0}=\mathcal{S}_e(\mathcal{G}_0, \hat{\mathcal{D}})$ is compressed prior link assumption after sparse link handling using $\mathcal{S}_e$. The $\Gamma_e$ is the edge pruning function that removes spurious links from multiple time lags at the post-processing stage.
We leave readers to Refs.~\cite{runge2019detecting, runge2020discovering, gerhardus2020high} for a comprehensive explanation of the working principle of the PCMCI algorithm, as we constrain the remaining discussion to our study.

\subsubsection{Causal Graph Generation}

We aim to capture the causality linkages on binary anomaly data, particularly the transition from being healthy with flag $0$ to experiencing an anomaly with flag $1$. 
The popular CI tests for categorical data, such as statistical \textit{G-squared}~\cite{bishop2007discrete} and information-theory \textit{conditional mutual information} tests~\cite{gerhardus2020high}, can not easily distinguish the significance of the anomaly transition behavior. 
These methods could result in incorrect causality inferred from the association influenced by the zeros instead of the ones. 

We propose $\hat{\mathcal{I}}$, an {anomaly-flag aware CI test} (ANAC), using a partial-correlation that only considers links with positive associations corresponding to anomaly occurrence. 
The PCMCI estimates GCM of a anomaly flag $\Lambda_i$ with a time-lagged function of the multivariate as:
\begin{equation}
\begin{aligned}
\Lambda_i &= \mathcal{F}({w_j(t-s), \Lambda_j(t-s)}), \\
&~\text{for}~s=0, \dots, \tau_{\text{max}}~\text{and}~j \in \mathbf{P A}_i,
\end{aligned}
\end{equation}
where the $w$ is the time-lagged causal weights; the $\tau_{\text{max}}$ is the maximum time-lag for causal inference; the $\mathbf{P A}_i$ is the set of the parent nodes or variables of $i$; the $\mathcal{F}$ is a binary GCM function. The CI test influences $w_j$ that positive $w_j$ indicates a positive correlation with $\Lambda_i$, increasing together during anomaly occurrence, and its value indicates strength of the correlation. 
Thus, we enhance the PCMCI~\cite{runge2020discovering} with simple but effective ANAC to detect anomaly occurrence causality by excluding causal time-lags with $\hat{\mathcal{I}}: w_j \leq 0$. 

The partial correlation is estimated through \textit{linear ordinary least squares regression} and positive \textit{Pearson's correlation} ($\rho$) CI test on the residuals. 
To test $X \upvDash Y | Z$, first $Z$ is regressed out from $X$ and $Y$ using the regression model:
\begin{equation}
\label{eq:rca__rho_reg_fit}
\begin{aligned}
& X=\beta_X Z+\epsilon_X, \\
& Y=\beta_Y Z+\epsilon_Y,
\end{aligned}
\end{equation}
where $\epsilon_X$ and $\epsilon_Y$ are residuals, and their independence is evaluated using Student's t-test on the $\rho\left(\epsilon_X, \epsilon_Y\right)$ to generate the $p_v$. Finally, the set of $j \notin \mathbf{P A}_i$ are excluded based on union of the significance CI test and non-positive correlation as: 
\begin{equation}
\begin{aligned}
&\hat{\mathcal{I}}: p_v > \alpha_p \cup w_j \leq 0. \\
\end{aligned}
\end{equation}

\subsubsection{Computation Cost of Graph Generation}

The computational cost of GCM for TS data can vary with the number of variables $N$, the data sample size $n$, the maximum time-lag $\tau_\text{max}$, and the choice of conditioning set size $d=|Z|$~\cite{runge2018causal}. 
The overall processing cost depends on the cost of the employed CI test $\mathcal{K}_\mathcal{I}$ and the number of CI tests $m$ that the CD algorithm executes to generate the GCM. The complexity of the CI test, $X \upvDash Y | Z$: $X$ independent of $Y$ conditioned on $Z$, is one major factor affecting the computational workload in a constraint-based algorithm~\cite{runge2018causal}. 
The cost of the GCM $\mathcal{K}(n, N,\tau_\text{max})$ can be formulated as:
\begin{equation}
   \mathcal{O} (\mathcal{K}) = \mathcal{O} \left( m.\mathcal{K}_\mathcal{I}(n, N,\tau_\text{max},d) \right), 
\end{equation}
where $\mathcal{K}_\mathcal{I}$ is the cost per a single CI test with conditioning set size up to $d$. 
The computational complexity of the PC algorithm has been
studied in Ref.~\cite{kalisch2007estimating}.
Although the exact estimation of the cost is difficult~\cite{li2009controlling}, we briefly formulate the cost as follows. 
The cost per $\mathcal{I}$, a regression test using partial correlation, is $\mathcal{O}(\mathcal{K}_\mathcal{I}) = \mathcal{O}(nd^2)$, as the main cost comes from the matrix manipulation with dimension $d \ll n$.
To simplify the mathematical formulation for $m$, let $\bar{N}=N.\tau_\text{max}$, approximating the number of variables in the temporal GCM. 
Considering CI tests without conditional set, i.e., $d=0$, each variable $x_i \in \mathcal{D}$ will conduct $\mathcal{I}(x_i, x_j),~\text{for}~ \forall j \neq i ~\text{and}~t=[-\tau_\text{max}, 0]$, resulting $\approx \bar{N}^2$ tests. 
Assuming the worst-case scenario where the PC tries all conditional sets $d=[0,\dots,\bar{N}-2]$ and each $d$ results in $\binom{\bar{N}-2}{d}$ selections, the $m$ becomes:
\begin{equation}
\label{eq:pc_m_ts_cost_max}
   m = \bar{N}^2 \sum _{d=0}^{\bar{N}-2}\binom{\bar{N}-2}{d} = \bar{N}^2 2^{\bar{N}-2}.
\end{equation}

For a bounded scenario with fixed $d$, the above Eq.~\eqref{eq:pc_m_ts_cost_max} reduces to:
\begin{equation}
\label{eq:pc_m_ts_cost_bounded}
   m = \bar{N}^2 \binom{\bar{N}-2}{d}.
\end{equation}
Thus, the final cost is approximated as:
\begin{equation}
\begin{aligned}
   \mathcal{O}(\mathcal{K}) = & \mathcal{O}\left(\bar{N}^2 \binom{\bar{N}-2}{d}nd^2\right) \\
   = & \mathcal{O}\left(N^2\tau_\text{max}^2 \binom{N\tau_\text{max}-2}{d}nd^2\right).
   \end{aligned}
\end{equation}

The value of $d$ depends strongly on the structure of the true GCM and the data characteristics.
However, the typical value of $d$ is small, ranging from 1 to 5, as sparsity is prevalent in most real-world graphs with few immediate parent nodes, and the majority of candidate links are eliminated quickly in the early steps before the CD algorithm reaches high-order conditioning sets.
However, even with the integrated sparsity-based pruning mechanism of PCMCI algorithms, binary data with long uniform signals presents a challenge, such as multicollinearity, where variables influence several others, which impacts the partial correlation CI test, causing the $d$ to grow and limit graph accuracy with many spurious links. 
Thus, optimizing computational efficiency by reducing the cost of the CT test function, $n \rightarrow n'$, and/or the number of CT tests, $N \rightarrow N'$, for a given $\tau_{\text{max}}$ is beneficial.

\subsubsection{Sparse Data Handling}

Data sparsity is inevitable in binary anomaly data, as anomalies occur rarely and may persist over time when they appear. For large $n$, the computation overhead is significant. We exploit the anomaly data sparsity to reduce the sample data size of $\Lambda$ using $\mathcal{S}_d (\Lambda)$ (see Algorithm \ref{alg:rca__sparse_handler} in \textupdate{Appendix \ref{sec:appendix_alg}}). The data compression lessen the computation to $\mathcal{O}\left(N^2\tau_\text{max}^2 \binom{N\tau_\text{max}-2}{d}n'd^2\right)$, where $n' \ll n$. 

The sparse data handling (SDH) $\mathcal{S}_d (\Lambda)$ compresses the long regions with uniform anomaly status (see \textupdateref{lines 3--9} in Algorithm \ref{alg:rca__sparse_handler}). Hence, anomaly CD can be better captured (by reducing multicollinearity due to the sparse regions) from the status transitions with considerably lower computation on the compressed flag data $\Lambda_c$. 
The algorithm preserves the first $l_m$ indices of the regions to capture the time-lag causality that ensures inference within time-adjacent anomaly occurrences while substantially reducing the size of the remaining sparse regions (see \textupdateref{lines 7--8}). 

We set the time length $l_m$ slightly greater than the causality searching $\tau_{\text{max}}$ to avoid false adjacency between anomalies at different time stamps after compression. The $l_m$ also regulates the contribution between collective trend drift anomalies and transient anomalies. Longer $l_m$ increases the influence of the collective anomalies on the causality estimation and vice versa.
\textupdate{Our method aims to reduce the $n$ to address the computational burden of the GCM learning process on large datasets.} 

\subsubsection{Sparse Graph Link Handling}

The CD may remain complex for high-dimensional data with large $N$ despite the reduction of the computation by $\mathcal{S}_d$~\cite{runge2020discovering}.
Hence, we incorporate a sparse link handling algorithm $\mathcal{S}_e$ that selects the initial prior links to reduce the searching space (see Algorithm \ref{alg:rca__sparse_link_assum_handling} in \textupdate{Appendix \ref{sec:appendix_alg}}). 
The method aims to ameliorate the computation by reducing the number of conditional tests between potential links.
The prior link assumption excludes self-lag links and links from non-coinciding variables on the binary anomaly data.

The assumption of self-lag exclusion $\mathbf{P A}_i \in \Lambda_j(t-s),~\forall j \neq i$ reduces the computation of the link search from $N^2 \tau_{\text{max}}^2 \rightarrow N(N-1)\tau_{\text{max}}^2$ (see \textupdateref{line 2} in Algorithm \ref{alg:rca__sparse_link_assum_handling}). 
The assumption of non-coinciding link exclusion $\Psi(\Lambda_i, \Lambda_j)<\alpha_\tau$ excludes links from variables with temporally non-coinciding flags and alleviates the computation further, $N(N-1) \tau_{\text{max}}^2 \rightarrow \sum_{i=1}^{N} N'_i\tau_{\text{max}}^2$, where $N'_i \leq N-1$ (see \textupdateref{lines 3--19}). The $\Psi(\Lambda_i, \Lambda_j)$ measures the the coincidence score $\lambda_\tau^{i,j}$ between the variables $\Lambda_i$ and $\Lambda_j$ and the threshold $\alpha_\tau$ determines whether their link should be considered during the GCM learning phase.

Algorithm \ref{alg:rca__sparse_link_assum_handling} starts by extending the anomaly flag regions by $\tau_{\text{max}}$ (see \textupdateref{line 4}), as the GCM search up to time lag of $\tau_{\text{max}}$, to calculate the coincidence score that measures the time-lagged co-occurrence among a pair of sensors (see \textupdateref{lines 7--12}). 
We first calculate the step-wise difference of binary flags $\Lambda(t)-\Lambda(t-1)$ to detect the state transition regions (see \textupdateref{line 3}). We employ a sliding rolling sum with a window size of $\tau_{\text{max}}$ and threshold the values above zero to create active windows (size of $\leq \tau_{\text{max}}$) around the positive flag transitions (see \textupdateref{lines 4--5}). 
The $\lambda_\tau^{i,j}$ is calculated based on the normalized score of the number of co-occurring flags between the active windows of a pair of sensors (see \textupdateref{lines 7--12}). A link between $i$ and $j$ is not considered during the GCM learning phase if the $\lambda_\tau^{i,j} < \alpha_\tau$, e.g., zero indicates no co-occurrence (see \textupdateref{lines 13--19}).

\subsubsection{Edge Pruning and Adjustment}
   
Multiple sensor variables can report anomaly flags simultaneously for continuous time ranges that might cause the PCMCI~\cite{runge2020discovering, gerhardus2020high} to generate PDAG, which includes spurious edges---multiple time-lags, bidirected, or cyclic links---when dealing with temporal anomaly data. 
We present a pruning algorithm $\Gamma_e$ as post-processing to overcome this challenge (see Algorithm \ref{alg:rca__undirected_edge_pruning} in \textupdate{Appendix \ref{sec:appendix_alg}}). The algorithm groups linked nodes and retains the lag with the highest weight or the earliest time-lag from the GCM.
The higher link weights indicate stronger causality, and older time-lags correspond to earlier causality that is temporally close to the transitioning edges. 
We employ a \textit{chi-square test} to direct the bidirected edges at $t=0$ when the correlation CI test falls short in detecting the directions because of lack of time delay causality and symmetric correlation score unable to direct the edge. 
The pruning excludes the spurious links to generate a curated DAG that is needed for building the inference BN for causality query in the later stages.
    
\subsection{Bayesian Network Modeling}  
    
We employ a BN to enable causality inference based on specific user query conditions beyond having a static graph skeleton. We utilize a \textit{Bayesian parameter estimator}~\cite{koller2009probabilistic} with a \textit{Bdeu} to learn the CPD parameters of the GCM.
The edges of the generated temporal GCM $\mathcal{G} \left( (\upsilon, \nu), \varepsilon (w, s) \right)$ have weight $w$ and time-lag $s$ attributes. To build a temporal causal BN, we reformulate as:
\begin{equation}
\label{eq:unroll_graph}
{\mathcal{G}} (\mathbf{V}, \mathbf{E}): (\upsilon, \nu) \rightarrow (\upsilon_s, \nu),~\varepsilon (w, s) \rightarrow \varepsilon(w_s),
\end{equation}
where source node $\upsilon$ with an edge $\varepsilon(w, s)$ is represented by new nodes $\upsilon_s$ and an edge $\varepsilon(w_s)$ for every active time-lag $s$. 

We prepare the data for training data for the BN by unrolling the TS data of the $\upsilon(t)$ (see Algorithm \ref{alg:rca__bn_training} in \textupdate{Appendix \ref{sec:appendix_alg}}). The approach is similar to the unrolling method adopted in TPC~\cite{biswas2022statistical} for temporal GCM. New variable names with $\upsilon_s$ tag for $\upsilon_s=\upsilon(t-s)$ are added by shifting the data ahead by the amount of time-lag $s$ (see \textupdateref{line 5--9}). Although there are also other tools for building TS BN, such as dynamic BNs, most are restricted to 2-time step temporal BN that requires only a unit time-lag links ($\tau_\text{max} \leq 1$) or existence of self-lag connections~\cite{murphy2002dynamic, sheidaei2022novel}. 

\section{Results and Discussion}
\label{sec:resultsanddiscussion}

This section presents the results of the performance evaluation of our \textsc{AnomalyCD} and benchmark state-of-the-art methods on the HCAL-RM and EasyVista TS datasets.

\subsection{Performance Metrics}

We will employ visual illustrations and several quantitative metrics to evaluate the proposed approaches.

We will compute \textit{structural hamming distance} (SHD) and \textit{area under the precision-recall curve} (APRC) to compare the discovered causal DAGs quantitatively with the ground truth GCMs. 
The SHD is a standard distance metric that compares acyclic graphs based on the counts of the edges that do not match~\cite{tsamardinos2006max, peters2015structural}. 
It counts two errors for a directed link with the reversed edge: for falsely directing the edge and for missing the correct edge.
\begin{equation}
\label{eq:rca__shd}
\begin{aligned}
 SHD (\mathcal{G}, \mathcal{Q}) \leftarrow  &\parallel \mathcal{G}  (\varepsilon(i, j)) \neq \mathcal{Q} (\varepsilon(i, j))\parallel \\
&~\text{for}~(i, j)~ \in~\mathbf{V}, 
\end{aligned}
\end{equation}
where $\mathbf{V}$ is the set of vertex nodes of the $\mathcal{G}$ and $\mathcal{Q}$ graphs, and $\parallel \parallel$ is a count function for the number of mismatched nodes between the $\mathcal{G}$ and $\mathcal{Q}$.

The APRC is a classification metric that evaluates performance using the area under the curve of the \textit{precision} (P) and \textit{recall} (R) coordinates. The P and R are defined as:
\begin{equation}
\label{eq:rca__clf_1}
\begin{aligned}
P = \frac{TP}{TP+FP},~\text{and}~
R = \frac{TP}{TP+FN},
\end{aligned}
\end{equation}
where the TP, FP, and FN stand for true positive, false positive, and false negative, respectively: 
the TP is the number of edges estimated with a correct direction; 
the FP is the number of edges that are in the estimated graph but not in the true graph;
the FN is the number of edges that are not in the estimated graph but in the true graph.

We will also utilize additional metrics, such as $\text{F}_\text{1}$-\textit{score}, \textit{false positive rate} (FPR), and \textit{undirected} SHD (SHDU), to compare the performance with benchmark CD approaches:
\begin{equation}
\label{eq:rca__clf_2}
    \begin{aligned}
 F_1 = &\frac {2 \times P \times R} {P + R}, ~
     FPR = \frac {RV + FP} {TN + FP}, ~\text{and} \\
     & SHDU = UE + UM + RV, 
    \end{aligned}
\end{equation}
where the SHDU penalizes edges with wrong directions once instead of twice like SHD in Eq.~\eqref{eq:rca__shd}~\cite{zhang2021gcastle}. 
The TN stands for true negative and is calculated as the number of edges that are neither in the estimated graph nor the true graph.  
The RV, UE, and UM denote the number of reversed edges, undirected extra edges, and undirected missing edges, respectively. 

\subsection{Causal Discovery on the HCAL}
\label{sec:rca__causal_hcal_data}

This section will discuss the performance of the proposed \textsc{AnomalyCD} using a use case HCAL-RBX system.
We will first discuss the performance of the online AD approach and then proceed to the anomaly CD.

\subsubsection{Online Anomaly Detection} 

Table~\ref{tbl:rca__od_settings_1min} provides the hyperparameter settings of the online algorithms incorporated in our ensemble AD approach. We set the anomaly thresholds slightly higher to reduce noise contamination and preserve the causal faithfulness assumption~\cite{spirtes2000causation, goudet2018learning, glymour2019review}.

\begin{table}[!htbp]
\caption{Hyperparameter settings for the online AD.}
\centering
\centering
    \begin{tabular}{@{}lp{4cm}@{}}
    \toprule
    \textbf{Algorithm}                        & \textbf{AD Settings}                                           \\
    \midrule
    $\Theta$ & $\alpha_\theta=10, ~w_\theta=5760$      \\ 
    $\mathcal{L}$      & $\alpha_\tau=20, ~p_\tau=5760,~k_\tau=5$         \\ 
    $\Phi$ & $\alpha_\phi=35, ~q_\phi=1440$                      \\
    \botrule
\end{tabular}
\label{tbl:rca__od_settings_1min}
\vspace*{-\baselineskip}
\end{table}

The LHC has undergone operations that result in distinct signal patterns on the sensors (see Fig.~\ref{fig:rca__lhc_ops_mask_aug_dec_2022}). We have thus utilized change point breaks (on 2022-09-27, 2022-10-19, and 2022-11-12) in which the AD reinitializes to deal with irrelevant data during operation interruptions. 

\begin{figure*}[!htbp]
\centering
\includegraphics[width=1\linewidth]{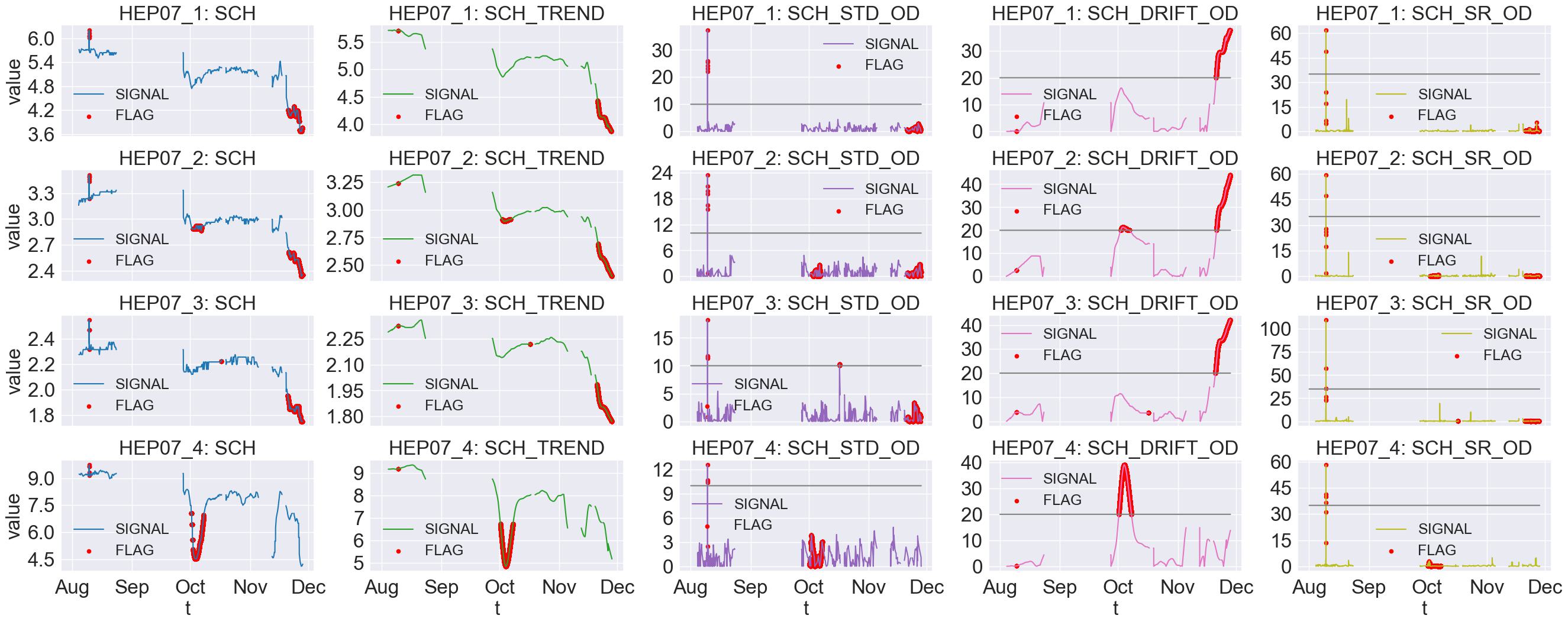}
\caption{Online AD on RBX-HEP07 SCH sensors. (Left to right) sensor signal, signal trend estimation, $\Lambda_\iota$ of $\mathcal{L}$, $\Lambda_\theta$ of $\Theta$, and $\Lambda_\phi$ of $\Phi$
}
\label{fig:rca__rm_sensor_data_HEP07_rm_all__01-08__31-11-2022_1min_SIPM__HUMIDITYS_F_ol_marked}
\vspace*{-\baselineskip}
\end{figure*}

\begin{figure*}[!htbp]
\centering
\includegraphics[width=1\linewidth]{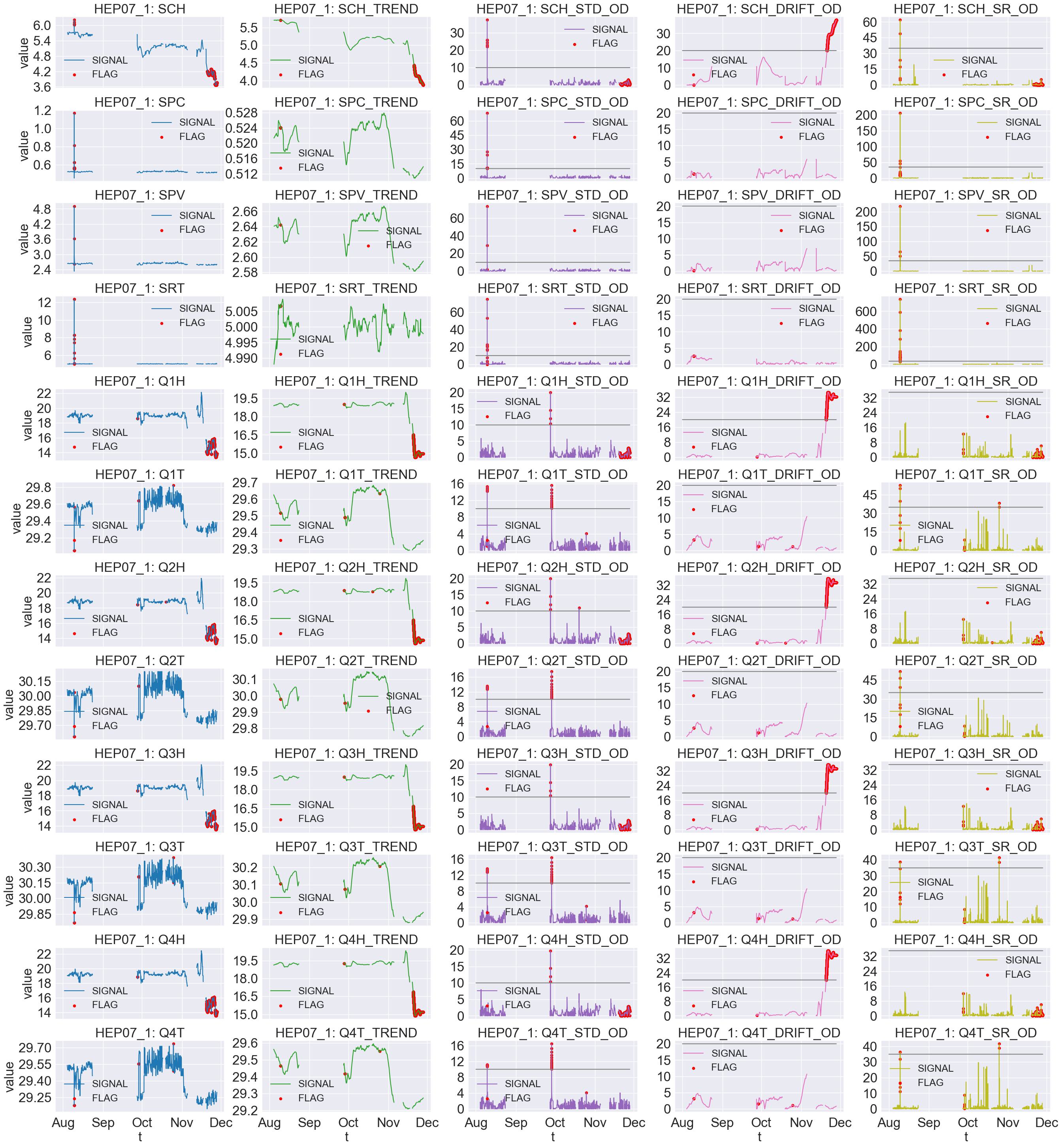}
\caption{Online temporal AD on the RBX-HEP07-RM-1 sensors
}
\label{fig:rca__rm_sensor_data_HEP07_1__01-08__31-11-2022_1min_ol_marked}
\vspace*{-\baselineskip}
\end{figure*}

Figure~\ref{fig:rca__rm_sensor_data_HEP07_rm_all__01-08__31-11-2022_1min_SIPM__HUMIDITYS_F_ol_marked} depicts the SCH sensors of the four RMs of HEP07 along with the detected transient and trend anomalies marked on the AD score signals of the AD algorithm. The sensors exhibit drifting trends that gradually deviate from the expected optimal values. 
Figure~\ref{fig:rca__rm_sensor_data_HEP07_1__01-08__31-11-2022_1min_ol_marked} illustrates all the sensors from RM-1 with marked anomalies. 
The humidity sensors have higher anomaly flag counts due to the detected trend drift anomaly. 

We have demonstrated the capability of our proposed online AD approach in detecting different types of anomalies on TS data with light computation overhead. The potential limitation of the approach is that it expects adequate healthy data samples, as it is challenging to detect anomalies without prior knowledge when the outlier is dominant in data that violates the rare anomaly occurrence assumption. In such cases, using trained AD models is recommended. 
Finding the optimal hyperparameters is another open challenge. Despite data normalization and standardization partially alleviating this challenge, hyperparameters tuning may require domain knowledge to capture the right data characteristics. 
However, hyperparameter adjustment on statistical models is lightweight and could be faster than tuning and retraining of DL models. 

\subsubsection{Graphical Causal Discovery} 

\begin{figure*}[!htbp]
\centering
\begin{subfigure}[]{1\linewidth}
\centering
\includegraphics[width=1\linewidth]{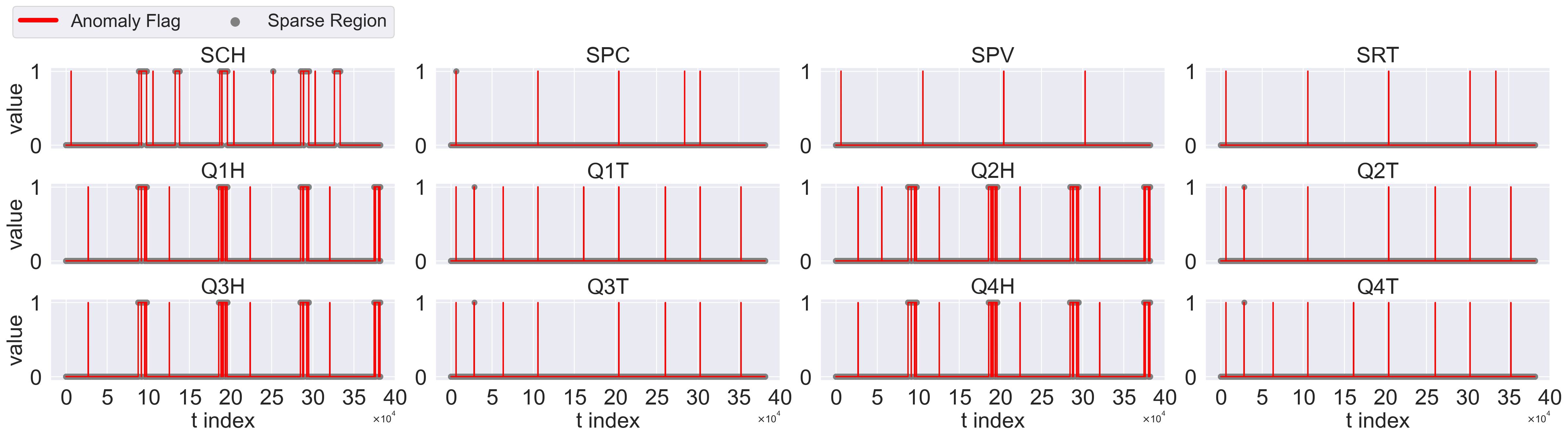}
\caption{}
\end{subfigure}
\begin{subfigure}[]{1\linewidth}
\centering
\includegraphics[width=1\linewidth]{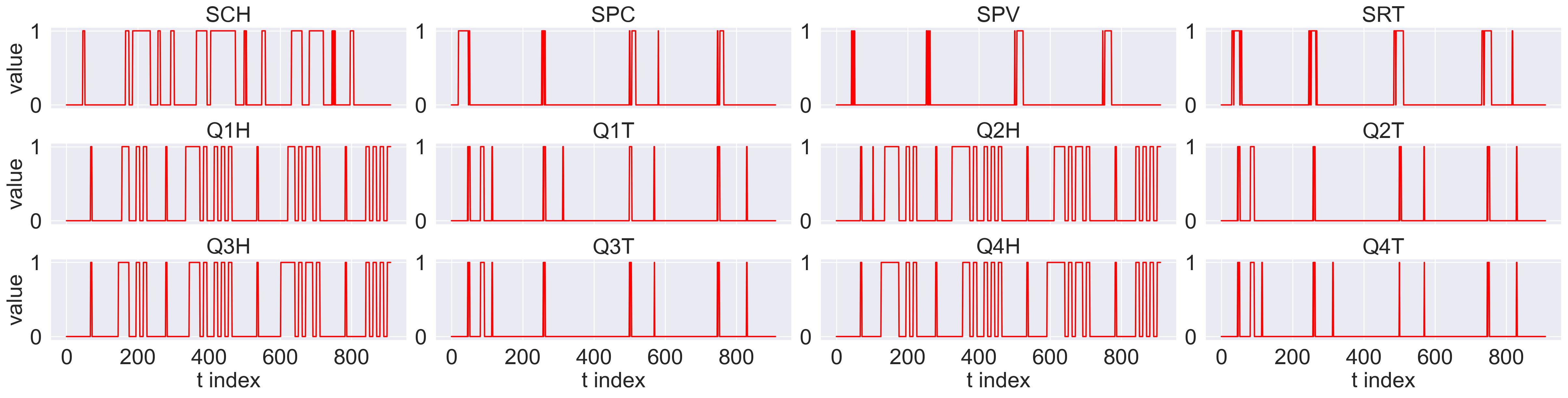}
\caption{}
\label{fig:rca__rm_sensor_data_HEP07_rm_all__01-08__31-11-2022_1min_sparse_compressed}
\end{subfigure}
\caption{Anomaly binary flag data from our proposed online AD approach on RBX-RM sensors:  
a) the raw anomaly data with approximately $4 \times 10^5$ samples and the sparse regions are annotated, and b) sparse compressed data through our $\mathcal{S}_d$ with $l_m=10$ and reducing the sample size to approximately $9 \times 10^2$
}
\label{fig:rca__rm_sensor_data_HEP07_rm_all__01-08__31-11-2022_1min_ol_flag_marked}
\vspace*{-\baselineskip}
\end{figure*}

We capture the GCM from the TS binary anomaly data generated by the online AD algorithm in the previous section (see Fig.~\ref{fig:rca__rm_sensor_data_HEP07_rm_all__01-08__31-11-2022_1min_ol_flag_marked}).
We employ data from multiple RMs to capture the global causality of the RM of the HE. 
We set the $\tau_{\text{max}}=5$ to search for temporal causality dependency at $[t - \tau_{\text{max}}, \dots, t]$ (equivalent to five minutes) and CI threshold $\alpha_p = 0.05$ for the PCMCI.

The sparse data handler $\mathcal{S}_d$ reduces the data by $99.8\%$ from approximately $400,000$ to $900$ samples. This greatly alleviates the computational cost of the GCM learning~\cite{runge2018causal, runge2020discovering}. The $\mathcal{S}_d$ took 8 seconds, and it squeezes the uniform regions to $l_m=10$ samples, i.e., twice of the $\tau_{\text{max}}=5$ of the CD (see Fig.~\ref{fig:rca__rm_sensor_data_HEP07_rm_all__01-08__31-11-2022_1min_sparse_compressed}).

\begin{figure}[]
\centering
\includegraphics[width=0.9\linewidth]{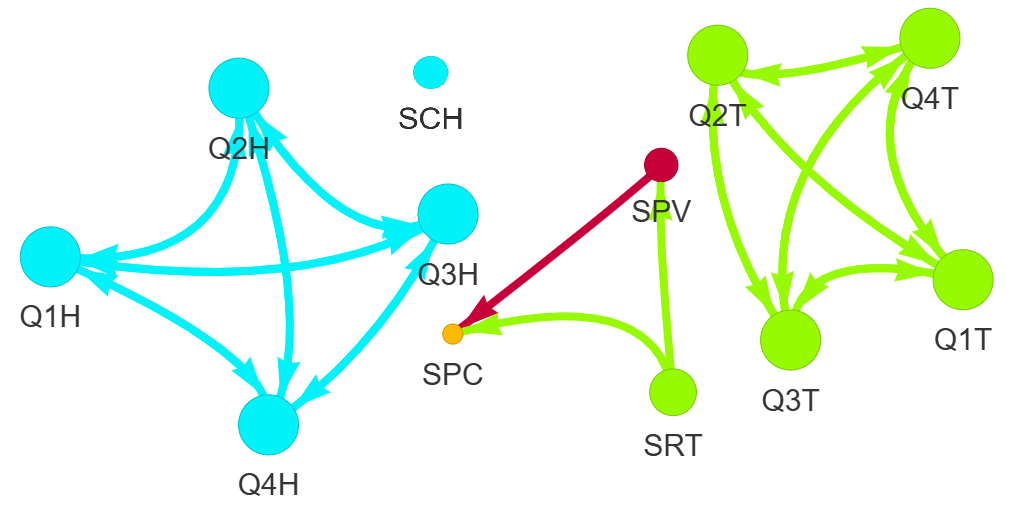}
\caption{Interconnection graph of the HCAL-RM system. We inferred the graph using domain knowledge of circuit connections, shared housing of the RM components, and the interconnection discovery result from Ref.~\cite{asres2024lightweight} 
}
\label{fig:rca__hcal_rm_ground_truth}
\vspace*{-\baselineskip}
\end{figure}

\begin{figure*}[!htbp]
\centering
\begin{subfigure}[]{0.55\linewidth}
\centering
\includegraphics[width=1\linewidth]{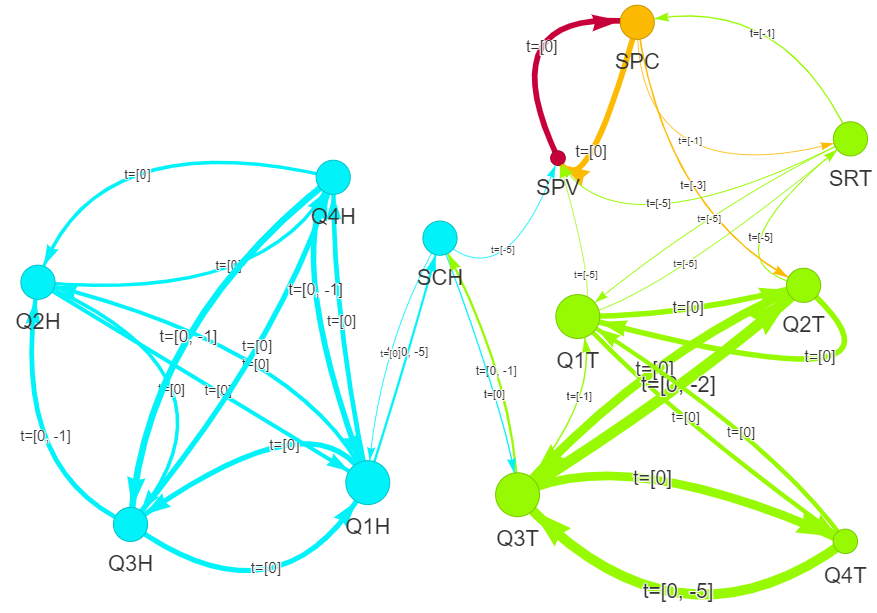}
\caption{}
\label{fig:rca__he_rm_HEP07_rm_all_rca_ts_1min_positive_0.05_white_5_}
\end{subfigure}
\begin{subfigure}[]{0.35\linewidth}
\centering
\includegraphics[width=1\linewidth]{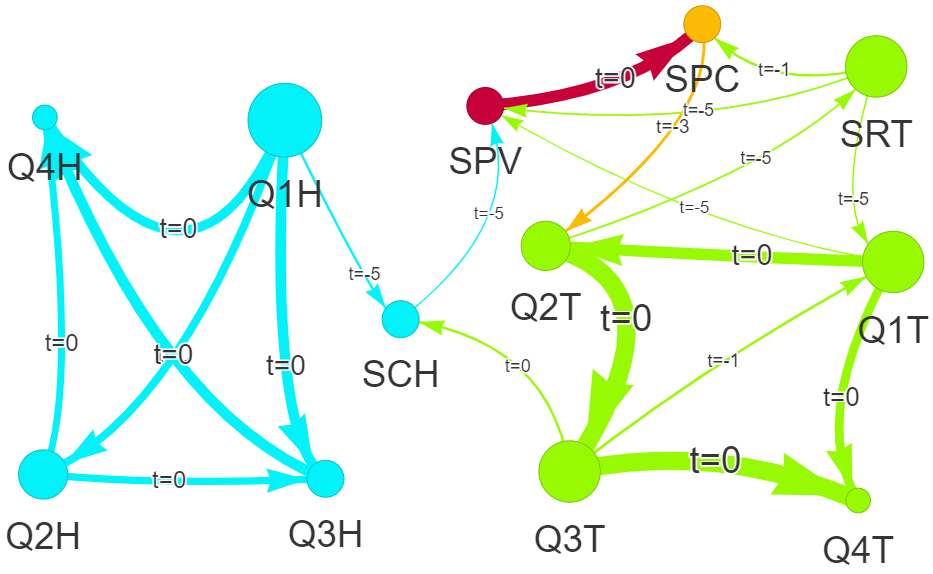}
\caption{}
\label{fig:rca__he_rm_HEP07_rm_all_rca_ts_bn_1min_positive_0.05_white_5_}
\end{subfigure}
\hfill
\begin{subfigure}[]{0.55\linewidth}
\centering
\includegraphics[width=1\linewidth]{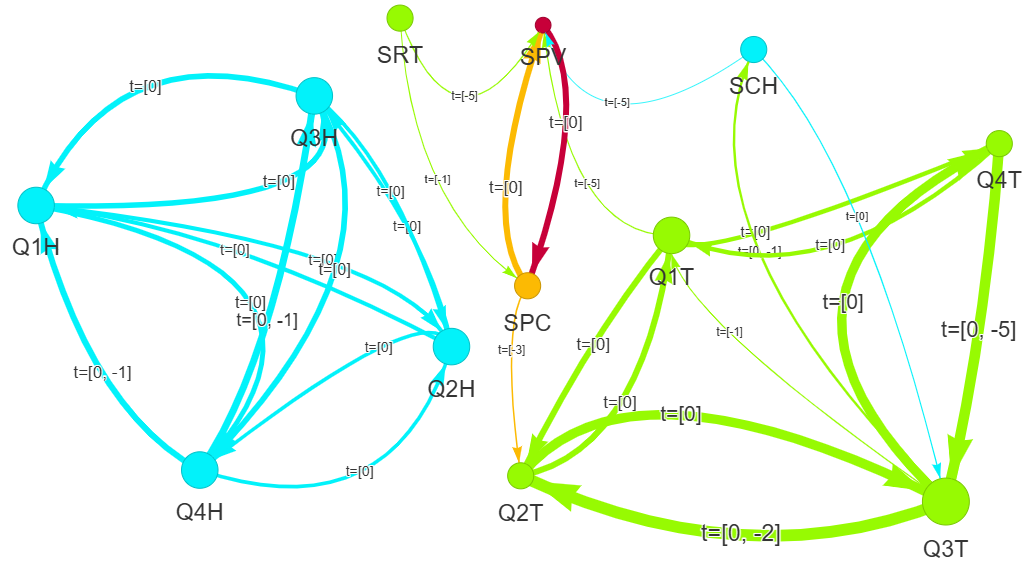}
\caption{}
\label{fig:rca__he_rm_HEP07_rm_all_rca_ts_1min_positive_0.05_white_5_with_sparselink_handling}
\end{subfigure}
\begin{subfigure}[]{0.35\linewidth}
\centering
\includegraphics[width=1\linewidth]{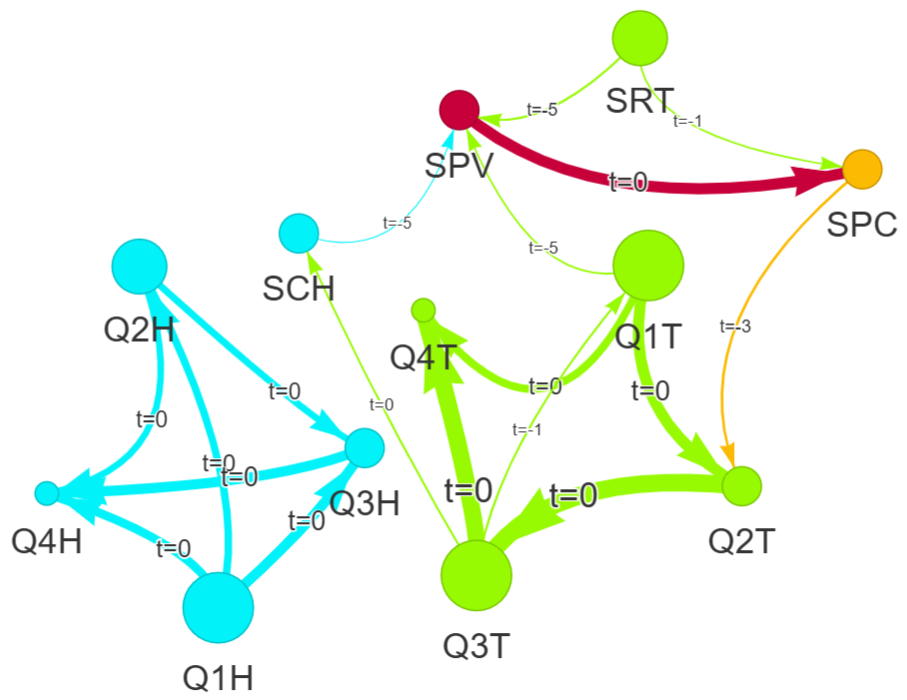}
\caption{}
\label{fig:rca__he_rm_HEP07_rm_all_rca_ts_bn_1min_positive_0.05_white_5_with_sparselink_handling}
\end{subfigure}

\caption{\textupdate{Temporal GCMs of HEP07-RM using $\tau_{\text{max}}=5$ on time-lag $t=[-\tau_{\text{max}}, 0]$: (top) with $\mathcal{S}_d$, and (bottom) with $\mathcal{S}_d + \mathcal{S}_e$. (a) and (c) are GCMs with PDAG before edge pruning, and (b) and (d) are GCMs with DAG after edge pruning
}}
\end{figure*}

We evaluate the discovered GCM by measuring the accuracy with respect to expected interconnection graph network depicted in Fig.~\ref{fig:rca__hcal_rm_ground_truth}.
Figure~\ref{fig:rca__he_rm_HEP07_rm_all_rca_ts_1min_positive_0.05_white_5_} illustrates the temporal GCM structure on the time-lag of $t=[-\tau_{\text{max}}, 0]$, captured by our \textsc{AnomalyCD} method before edge link pruning. 
The GCM shows expected interconnection among the sensors: the clustering of environmental temperature and humidity sensors, and the link between the temperature regulator Peltier voltage and current, and the corresponding temperature sensors. Several bidirected edges with multiple time-lags are also present in the network. 
Figure~\ref{fig:rca__he_rm_HEP07_rm_all_rca_ts_bn_1min_positive_0.05_white_5_} shows the final temporal DAG after applying Algorithm~\ref{alg:rca__undirected_edge_pruning} for the pruning. 
An interesting time-lagged causality link can be observed, where a temperature anomaly in SRT leads to anomalies in SPV and SPC; this is because the Peltier regulator increases SPV and SPC to respond to SRT's upsurge.

\textupdate{Table~\ref{tbl:rca__hcal_rm__sparse_handler_prec} presents quantitative performance of the graphical CD for the HCAL-RBX. The PDAG from the PCMCI algorithm, with spurious links at multiple time lags, leads to higher inaccuracy and lower performance. Our AnomalyCD has improved the CD substantially with a relative gain of $26.2\%$ in F$_1$, $74.9\%$ in FPR, $17.1\%$ in APRC, and $70.6\%$ in SHDU. The ANAC delivers improvement of around $16\%$ in F$_1$, $44\%\text{--}56\%$ in FPR, $9\%\text{--}12\%$ in APRC, and $41\%\text{--}53\%$ in SHDU. The impact of edge pruning $\Gamma_e$ remains limited when $\mathcal{S}_e$ is employed, as most of the potential irrelevant edges are excluded by the $\mathcal{S}_e$.
Further analysis of the CD over varying lengths of $l_m$ is provided in the Appendix~\ref{sec:rca__causal_hcal_data_multi_lm}. 
}

\begin{table*}[!htbp]
\caption{Ablation study on \textsc{AnomalyCD} graphical CD using HCAL-RM dataset. The \textsc{AnomalyCD}: PCMCI + SDLH + ANAC + Edge Pruning. 
The SDH and SDLH denote $\mathcal{S}_d$ and $\mathcal{S}_d + \mathcal{S}_e$, respectively. 
The PT is the processing time in seconds}
\centering
\resizebox{1\linewidth}{!}{
\begin{tabular}{@{}llllll@{}}
\toprule
\textbf{Method}  & \textbf{F$_1$}$\uparrow$ &  \textbf{FPR}$\downarrow$ &  \textbf{APRC}$\uparrow$ & \textbf{SHDU}$\downarrow$ & \textbf{PT} (secs)$\downarrow$ \\
\midrule
\textsc{AnomalyCD}: PCMCI + SDH                     & 0.667 (--)              & 0.410 (--)               & 0.730 (--)              & 17 (--)              & 9.148 (--)               \\
\textsc{AnomalyCD}: PCMCI + SDH + Edge Pruning         & 0.696 (4.3\%)           & 0.410 (--)               & 0.741 (1.5\%)           & 17 (--)              & 9.190 (0.5\%)            \\
\textsc{AnomalyCD}: PCMCI + SDH + ANAC              & 0.774 (16.0\%)          & 0.180 (-56.1\%)          & 0.798 (9.3\%)           & 8 (-52.9\%)          & 8.966 (-2.0\%)           \\
\textsc{AnomalyCD}: PCMCI + SDH + ANAC + Edge Pruning  & 0.800 (19.9\%)          & 0.180 (-56.1\%)          & 0.818 (12.1\%)          & 8 (-52.9\%)          & 8.997 (-1.7\%)           \\
\textsc{AnomalyCD}: PCMCI + SDLH                    & 0.725 (8.7\%)           & 0.282 (-31.2\%)          & 0.768 (5.2\%)           & 12 (-29.4\%)         & 6.641 (-27.4\%)          \\
\textsc{AnomalyCD}: PCMCI + SDLH + Edge Pruning        & 0.719 (7.8\%)           & 0.308 (-24.9\%)          & 0.751 (2.9\%)           & 13 (-23.5\%)         & 6.680 (-27.0\%)          \\
\textsc{AnomalyCD}: PCMCI + SDLH + ANAC             & \textbf{0.842 (26.2\%)} & \textbf{0.103 (-74.9\%)} & \textbf{0.855 (17.1\%)} & \textbf{5 (-70.6\%)} & \textbf{6.553 (-28.4\%)} \\
\textsc{AnomalyCD}: PCMCI + SDLH + ANAC + Edge Pruning & \textbf{0.842 (26.2\%)} & \textbf{0.103 (-74.9\%)} & \textbf{0.855 (17.1\%)} & \textbf{5 (-70.6\%)} & 6.588 (-28.0\%)         \\

\botrule
\end{tabular}
}
\par  \footnotesize  The \textbf{black bold font} is the best score. The downarrow ($\downarrow$) means lower is better, and vice versa for uparrow ($\uparrow$). The percentages are calculated relative to the performance of the PCMCI + SDH. \par
\label{tbl:rca__hcal_rm__sparse_handler_prec}
\vspace*{-\baselineskip}
\end{table*}

\begin{figure*}[h]
\centering
\includegraphics[width=0.7\linewidth]{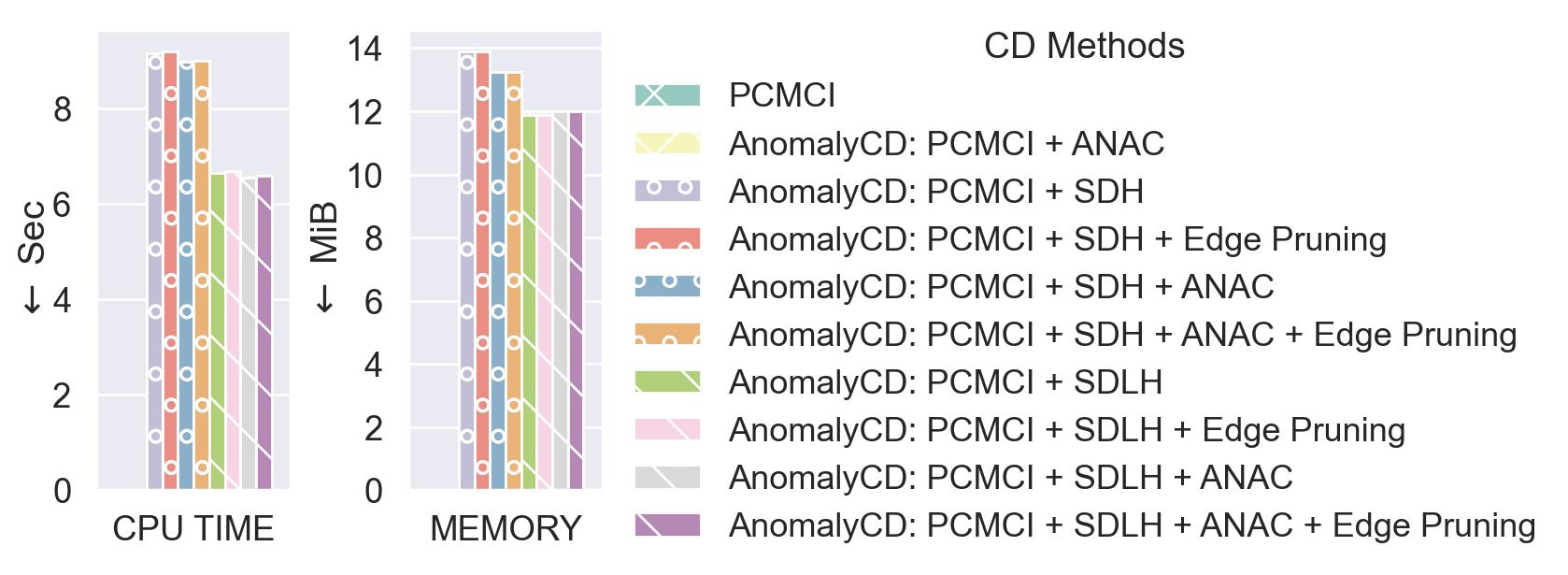}
\caption{Computational cost analysis on GCM learning on the HCAl-RM dataset. Our SDH $\mathcal{S}_d$ reduces the data from 400 thousand (the PCMCI failed with extremely slow processing) to 900 ($99.8\%\downarrow$) samples. The CD with $\mathcal{S}_d$  consumes approximately 14~MB and executes in 9~seconds. The $\mathcal{S}_e$ of the SDLH reduces the initial graph edges from 852 to 360 ($57.8\% \downarrow$), resulting in a gain of $14\%$ and $27\%$ in memory and process time, respectively, as compared to the $\mathcal{S}_d$
}
\label{fig:cc_CMS_HCAL}
\vspace*{-\baselineskip}
\end{figure*}

Figure~\ref{fig:cc_CMS_HCAL} portrays the computational cost of the CD using CPU processing time and incremental memory usage during the GCM learning phase. 
We conducted our experiment on 13th Gen Intel(R) Core(TM) i9-13980HX~@~2.2GHz with 64GB RAM. 
The PCMCI CD discovery on the raw data with over 400 thousand samples failed due to extremely slow processing. The \textsc{AnomalyCD} with SDH $\mathcal{S}_d$ consumes approximately 14~MB and executes in 9 seconds, with $99.8\%$ compression in data size. The SDLH ($\mathcal{S}_d+\mathcal{S}_e$) contributes a further boost in memory and processing time by $14\%$ and $27\%$ as compared to the $\mathcal{S}_d$, respectively, by reducing the prior link assumption edges by $57.8\%$. We found the cost of $\Gamma_e$ marginal with less than $0.5\%$ increment in processing time. 

\subsubsection{Bayesian Causality Inference} 

We trained our BN model, the query engine for temporal GCM inference shown in Fig.~\ref{fig:rca__online_rca-rca_online_main}, using unrolled TS data and the captured anomaly DAG skeleton.  
The causality queries on the BN include conditional probability (CP) inference, allowing users to query for the marginal distribution of anomaly occurrence. The inference enables estimating the causal effect between two nodes given observed anomaly evidence on the others. We employ the \textit{variable elimination} algorithm, an exact inference technique for solving Bayes' equations, to estimate CPs over a subset of variables from the probabilistic graphical BN models: 
\begin{equation}
    C_i = \wp (\mathcal{M}_b, x_i, S),
\end{equation}
where $C_i$ is the CP of the $x_i \in \{0, 1\}$ sensor for its anomaly flag state $s_i=1$ given the observed evidence on other sensors with states of $S=\{s_j \in \{0, 1\}, ~ \forall j \neq i \}$. The $\wp$ is the inference engine with the BN model $\mathcal{M}_b$.

Table~\ref{tbl:rca__sample_bn_result_1} presents results of anomaly causality inference queried from the $\mathcal{M}_b$. We quantify the causality by calculating the CP of the cause or affected sensors. 
The CP of anomaly occurrence for the Q1T sensor increases from $0.05$ with no other evidence to above $0.90$ with the evidence of a detected anomaly flag on the related Q[2-4]T sensors at a time-lag $t=0$. The CP of an anomaly on the Q1H sensor increases from $0.26$ to above $0.85$ when there is evidence of a detected anomaly flag on the Q[2-4]H sensors at a time-lag of $t=0$. 
The Q1H has a higher CP with no other evidence scenario due to trend drifts. We notice that the few sample differences during drift detection across the Q[1-4]H sensors lower the causality dependency strength.
The CP of SPC increases to $0.45$ when SPV has an anomaly at $t=0$. The SRT is causal to the SPC at a time-lag $t=-1$ directly and at $t=-5$ through SPV, but with uneven edge strength of $0.32$ and $0.15$, respectively. The CP rises to $0.95$ on the SPC when anomalies are detected on both the SPV and SRT, indicating the CP is influenced by edge weight strength from multiple causal nodes.

The BN inference has generally produced CPs that are aligned with the link strength of the causal DAG, depicted in Fig.~\ref{fig:rca__he_rm_HEP07_rm_all_rca_ts_bn_1min_positive_0.05_white_5_}. But, caution should be taken when approaching the BN causality interpretation: 1) edges between Q[1-4]T and Q[1-4]H (sensors from the closely placed QIE cards) indicate the BN inference depicts the anomaly relationship rather than the causality between the sensors, and 2) CP, whether it is causal or influenced by observed evidence, must be explained with link edge direction; for example, the CP increase in the SPC is due to the anomaly on the causal SPV node, whereas the increase on the SPV is from its influenced on the SPC node. 

\begin{table*}[] \small
\centering
\caption{Anomaly conditional probability $\mathcal{P}$ based on Bayesian causality inference using pruned GCMs. 
The $A=1$ and $B=1$ denote the target and observed sensors with active anomaly flags
}
\resizebox{1\linewidth}{!}{
\begin{tabular}{@{}llll@{}}
\toprule
\textbf{Target Sensor ($\mathbf{A=1}$)}     & \textbf{Observed Sensors ($\mathbf{B=1}$)}     & $\mathbf{\mathcal{P}_\text{SDH}(A=1 \mid B=1)}$ & $\mathbf{\mathcal{P}_\text{SDLH}(A=1 \mid B=1)}$ \\
\midrule
Q1T ($t=0$)                  & -                          & 0.051            & 0.052             \\
Q1T ($t=0$)                  & Q2T ($t=0$)                & 0.904          & 0.906             \\
Q1T ($t=0$)                  & Q{[}2-4{]}T ($t=0$)        & \textbf{0.927}          & \textbf{0.928}             \\ \hline
Q1H ($t=0$)                  & -                          & 0.262          & 0.262             \\
Q1H ($t=0$)                  & Q2H ($t=0$)                & 0.846          & 0.846             \\
Q1H ($t=0$)                  & Q{[}2-4{]}H ($t=0$)        & \textbf{0.913}          & \textbf{0.913}             \\ \hline
SPC ($t=0$)                  & -                          & 0.071          & 0.071             \\
SPC ($t=0$)                  & SPV ($t=0$)                & 0.456          & 0.456             \\
SPC ($t=0$)                  & SRT ($t=-1$)               & 0.321          & 0.321             \\
SPC ($t=0$)                  & SPV ($t=0$) and SRT ($t=-1$)& \textbf{0.947}          & \textbf{0.947}             \\ \hline
SPV ($t=0$)                  & -                          & 0.057          & 0.057             \\
SPV ($t=0$)                  & SPC ($t=0$)                & \textbf{0.366}          & \textbf{0.366}             \\     

\botrule
\end{tabular}
}
\label{tbl:rca__sample_bn_result_1}
\vspace*{-0.5\baselineskip}
\end{table*}

\subsection{Causal Discovery on EasyVista}
\label{sec:rca__causal_public_data}

\begin{figure}[!htbp]
\centering
\includegraphics[width=0.8\linewidth]{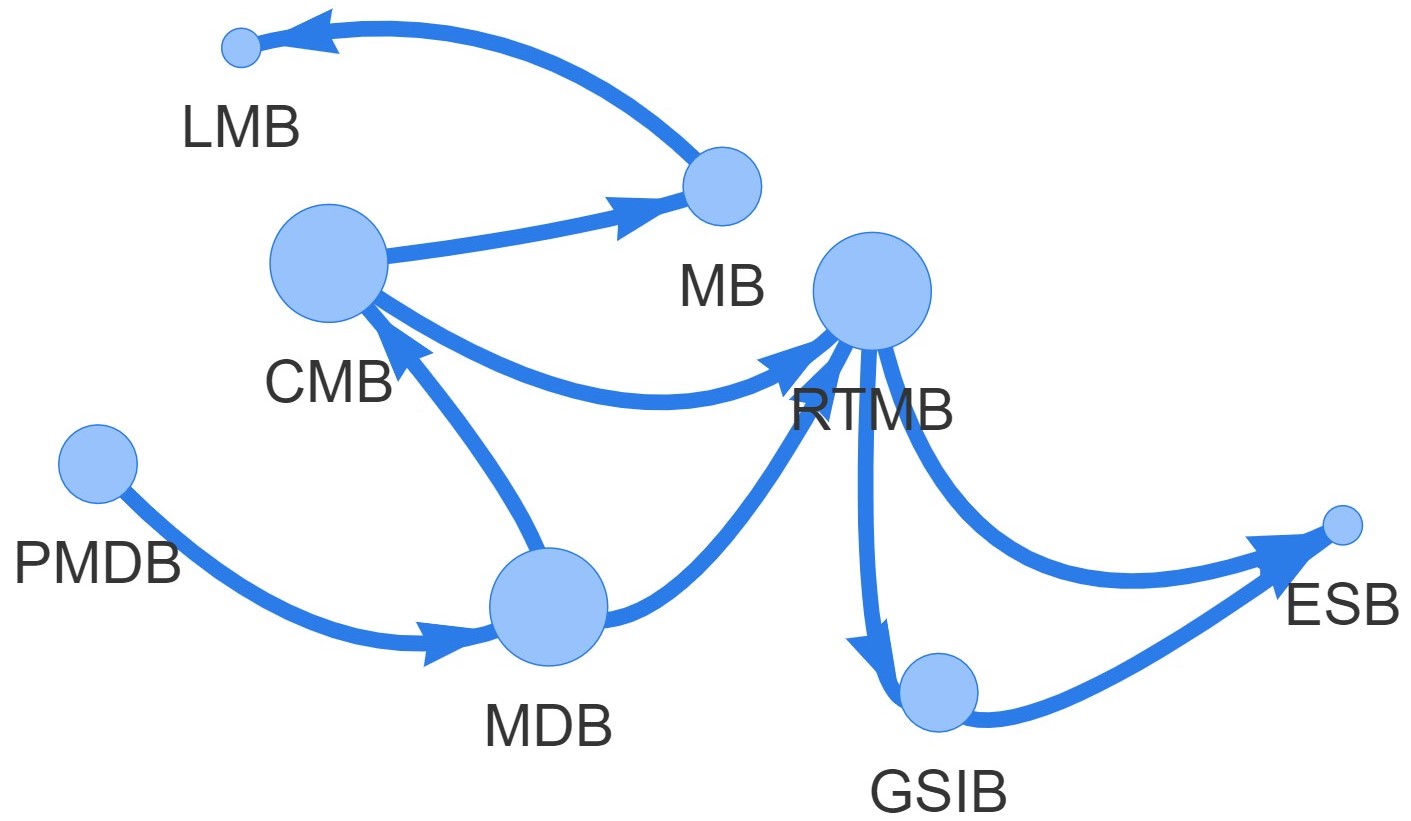}
\caption{Causal graph of EasyVista's monitoring system during normal operation
}
\label{fig:rca__easyrca_ground_truth}
\vspace*{-\baselineskip}
\end{figure}

We have tested our \textsc{AnomalyCD} approach on a publicly available EasyVista dataset~\cite{EasyRCA2023git} (see Section~\ref{sec:dataset_easyvista}). 
The dataset have been utilized the dataset for RCA study using a prior known causal graph network of the sensors~\cite{assaad2023root}(see Fig.~\ref{fig:rca__easyrca_ground_truth}).

\begin{figure*}[!htbp]
\centering
\begin{subfigure}[]{1\linewidth}
\centering
\includegraphics[width=1\linewidth]{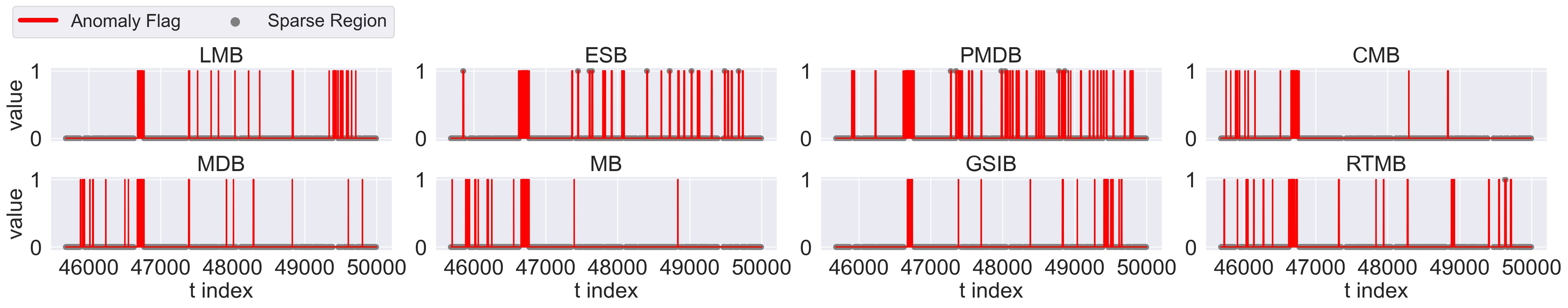}
\caption{}
\end{subfigure}
\begin{subfigure}[]{1\linewidth}
\centering
\includegraphics[width=1\linewidth]{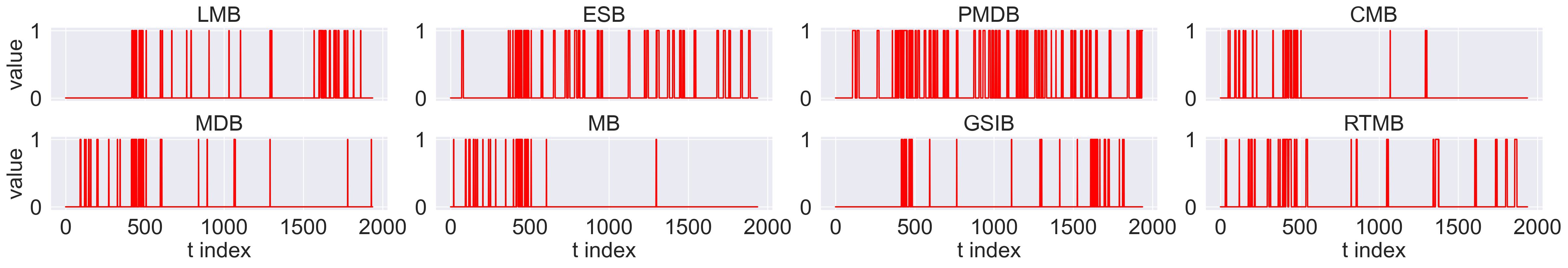}
\caption{}
\label{fig:EasyVista_od_sparse_compressed}
\end{subfigure}
\caption{The generated TS anomaly-flag data using our online AD on the EasyVista sensors:  
a) the raw anomaly data with approximately $4,300$ samples and the sparse regions are annotated, and b) compressed data through our $\mathcal{S}_d$ with $l_m=10$ and reducing the sample size by approximately $55\%$
}
\label{fig:rca__easyrca_idx_ol_flag_and_count}
\vspace*{-\baselineskip}
\end{figure*}

Figure~\ref{fig:EasyVista_od_sparse_compressed} presents generated the anomaly flags data using our online AD approach.
We have utilized a low threshold $\alpha_\phi=2$ to detect more outliers beyond the main collective anomaly at indices $46,683$--$46,783$ to improve the CD learning.
The EasyVista experts consider the PMDB and ESB  variables, with the highest anomaly flags, as the root causes of the anomalies~\cite{EasyRCA2023git,assaad2023root}. 

We utilize the causal graph of the normal operation, given in Fig.~\ref{fig:rca__easyrca_ground_truth}, as a reference graph to evaluate the accuracy of the estimated GCMs.
Table~\ref{tbl:rca__easyrca_compare_cd_methods} presents some of the popular benchmarking CD methods in the literature. We have preserved the undirected edges as bidirected for a fair comparison among the methods, as some of the methods generate PDAG. Since the ground truth graph does not contain temporal information, we convert the temporal GCM results into a summary GCM for the PCMCI temporal CD, i.e., we aggregate the time lag attributes of the edges into $t=0$. 
We utilize partial correlation for the CI testing, and a threshold $\alpha_p=0.05$ for the constraint-based methods.
We employ {BIC} and {BDeu} scores~\cite{carvalho2009scoring} for the score-based algorithms.

\begin{table}[] \small
\caption{Graphical causal discovery methods}
\centering
\begin{tabular}{@{}lp{4.5cm}@{}}
\toprule
\textbf{Method Category} & \textbf{List of Methods}  \\ 
\midrule
Constraint-based & PC~\cite{colombo2014order}, GS~\cite{margaritis2003learning}, IAMP~\cite{tsamardinos2003algorithms}, MMPC~\cite{tsamardinos2003time}, PCMCI~\cite{runge2020discovering} \\ 
Score-based & HC~\cite{koller2009probabilistic}, GES~\cite{chickering2002optimal} \\ 
Hybrid-based & MMHC~\cite{tsamardinos2006max} \\
Function-based & Direct-LiNGAM~\cite{shimizu2011directlingam}, ICA-LiNGAM~\cite{shimizu2006linear} \\ 
Gradient-based & GraN-DAG~\cite{lachapelle2019gradient}, GOLEM~\cite{ng2020role}, GAE~\cite{ng2019graph}, RL-BIC~\cite{zhu2019causal}, CORL~\cite{wang2021ordering} \\ 
\botrule
\end{tabular}
\label{tbl:rca__easyrca_compare_cd_methods}
\vspace*{-\baselineskip}
\end{table}

Table~\ref{tbl:rca__easyrca_with_sparse_handler} provide the CD performance on the raw and compressed anomaly data before and after $\mathcal{S}_d$, respectively. 
The $\mathcal{S}_d$ reduces the data size to $1,900$ samples ($55\%\downarrow$), which lowers computation time by $8$--$10$ times. It increases the accuracy of the causal graph by improving the precision (reducing false edges), which is demonstrated by the higher F$_1$, and lower FPR and SHDU. The score-based HC~\cite{koller2009probabilistic} and its hybrid MMHC~\cite{tsamardinos2006max} algorithms have limited accuracy gain on the compressed data. 
Only the graph autoencoder (GAE)~\cite{ng2019graph} has succeeded from the DL models.  The RL-BIC~\cite{zhu2019causal} and CORL~\cite{wang2021ordering} employ reinforcement learning and have missed all relevant edges.
Our $\mathcal{S}_d$ achieves an average relative improvement of $18.31\%$, $22.05\%$, and $15.45\%$ in the F$_1$, FPR, and SHDU, respectively, compared to raw data.

\begin{table*}[]  
\caption{Causal graph learning on the EasyVista dataset without and with (*) sparse handling
}
\centering
\resizebox{1\linewidth}{!}{
\begin{tabular}{@{}llllllllll@{}}
\toprule
\textbf{Method}                                                    & \textbf{Method Category}                                                    & \textbf{F$_1$}$\uparrow$ & \textbf{F$_1$}$^*\uparrow$ & \textbf{FPR}$\downarrow$ & \textbf{FPR}$^*\downarrow$ & \textbf{APRC}$\uparrow$ & \textbf{APRC}$^*\uparrow$ & \textbf{SHDU}$\downarrow$ & \textbf{SHDU}$^*\downarrow$ \\
\midrule
PC~\cite{colombo2014order} & Constraint-based                  & 0.154                                     & 0.333                                     & 0.790                                     & 0.474                                     & 0.395                                    & 0.225                                    & 18                                         & 12                                         \\
GS~\cite{margaritis2003learning} & Constraint-based            & 0.291                                     & 0.267                                     & 0.790                                     & 0.632                                     & 0.357                                    & \textbf{0.539}                           & 16                                         & 15                                         \\
IAMB~\cite{tsamardinos2003algorithms} & Constraint-based       & 0.291                                     & 0.267                                     & 0.790                                     & 0.632                                     & 0.357                                    & \textbf{0.539}                           & 16                                         & 15                                         \\
MMPC~\cite{tsamardinos2003time} & Constraint-based             & 0.291                                     & 0.308                                     & 0.790                                     & 0.474                                     & 0.457                                    & \textbf{0.539}                           & 16                                         & 12                                         \\
HC-BIC~\cite{koller2009probabilistic}& Score-based   & 0.191                                     & 0.100                                     & 0.526                                     & 0.526                                     & 0.164                                    & 0.249                                    & 13                                         & 15                                         \\
HC-BDeu~\cite{koller2009probabilistic}& Score-based  & 0.296                                     & 0.148                                     & 0.737                                     & 0.842                                     & 0.221                                    & 0.372                                    & 17                                         & 19                                         \\
GES-BIC~\cite{chickering2002optimal}& Score-based    & 0.235                                     & 0.323                                     & 0.895                                     & 0.632                                     & 0.423                                    & 0.341                                    & 18                                         & 14                                         \\
GES-BDeu~\cite{chickering2002optimal}& Score-based   & 0.267                                     & 0.267                                     & 0.526                                     & 0.526                                     & 0.357                                    & 0.357                                    & 15                                         & 15                                         \\
MMHC~\cite{tsamardinos2006max}& Hybrid-based               & 0.200                                     & 0.105                                     & 0.474                                     & 0.474                                     & 0.168                                    & 0.257                                    & \textbf{12}                                & 14                                         \\
Direct-LiNGAM~\cite{shimizu2011directlingam}& Function-based & 0.242                                     & 0.222                                     & 1.050                                     & 0.790                                     & 0.297                                    & 0.345                                    & 21                                         & 17                                         \\
ICA-LiNGAM~\cite{shimizu2006linear} & Function-based         & 0.235                                     & 0.308                                     & 1.105                                     & 0.684                                     & 0.379                                    & 0.341                                    & 22                                         & 16                                         \\
GOLEM~\cite{ng2020role}   & Gradient-based                   & 0.191                                     & 0.191                                     & 0.526                                     & 0.526                                     & 0.249                                    & 0.249                                    & 16                                         & 16                                         \\
GraN-DAG~\cite{lachapelle2019gradient}& Gradient-based       & 0.222                                     & 0.333                                     & \textbf{0.368}                            & \textbf{0.316}                            & 0.380                                    & 0.277                                    & 13                                         & 12                                         \\
GAE~\cite{ng2019graph} & Gradient-based                      & 0.231                                     & 0.250                                     & 0.421                                     & 0.421                                     & 0.314                                    & 0.302                                    & 14                                         & 13                                         \\
PCMCI~\cite{runge2020discovering} & Constraint-based           & 0.302                                     & 0.316                                     & 0.895                                     & 0.632                                     & 0.460                                    & 0.543                                    & 17                                         & 13                                         \\      
\textsc{AnomalyCD} (ours)  & Constraint-based                       & \textbf{0.304}                            & \textbf{0.364}                            & 0.684                                     & 0.474                                     & \textbf{0.499}                           & 0.482                                    & 17                                         & \textbf{10}                      \\       
\botrule
\end{tabular}
}
\par \footnotesize The \textbf{black bold font} is the best score. Downarrow ($\downarrow$) means lower is better, and vice versa for uparrow ($\uparrow$). \par
\label{tbl:rca__easyrca_with_sparse_handler}
\vspace*{-\baselineskip}
\end{table*}

\begin{figure*}[h]
\centering
\begin{subfigure}[]{0.35\linewidth}
\centering
\includegraphics[width=1\linewidth]{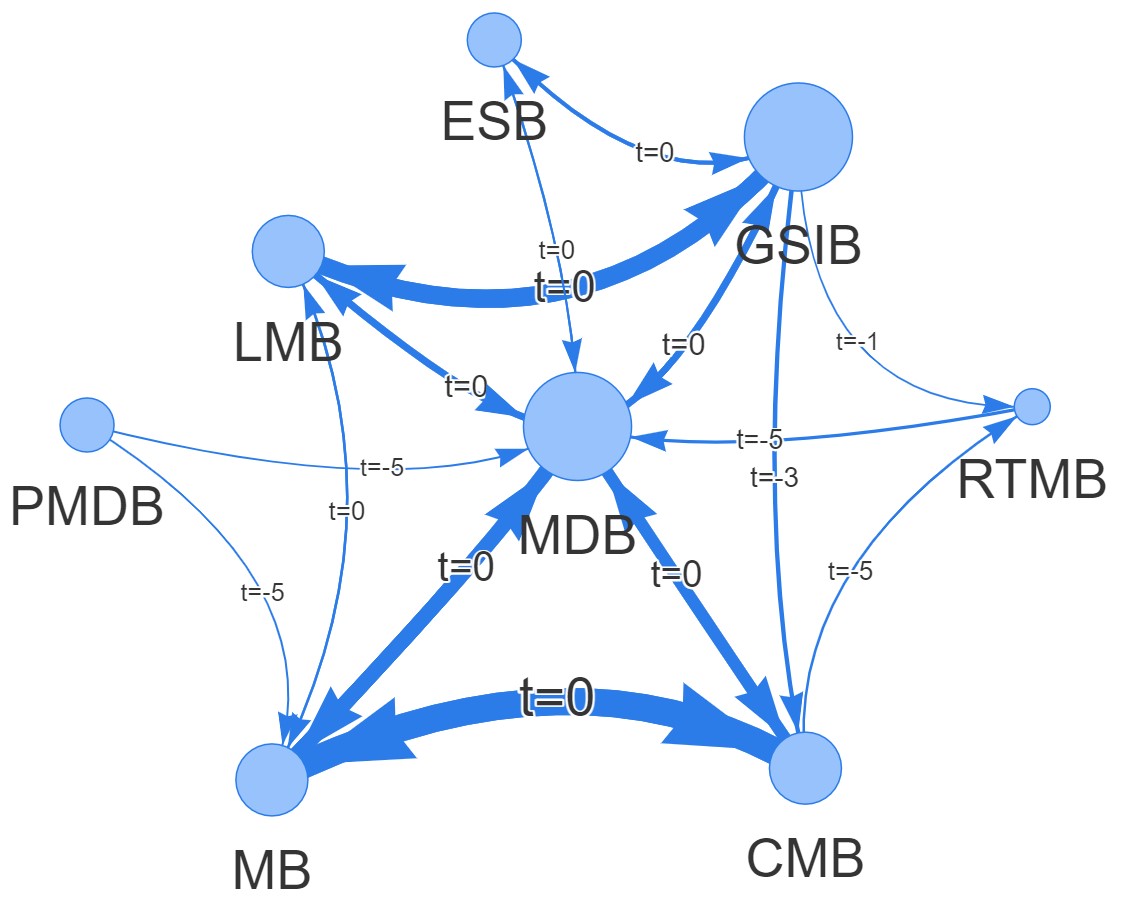}
\caption{}
\label{rca__easyrca_rca_ts_idx_positive_white_5_prune_tlag_first_undirected}
\end{subfigure}
\begin{subfigure}[]{0.3\linewidth}
\centering
\includegraphics[width=1\linewidth]{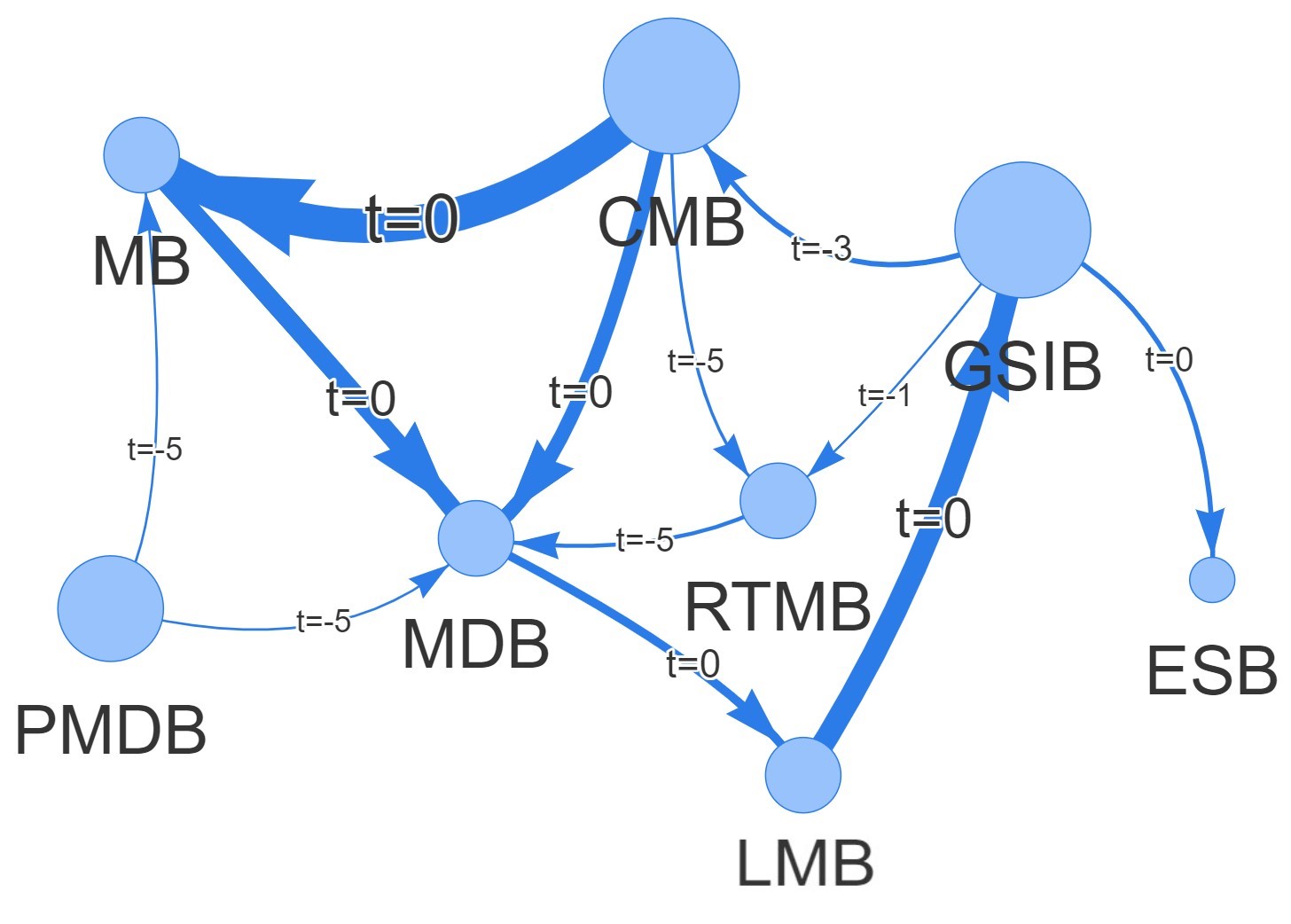}
\caption{}
\label{fig:rca__easyrca_rca_ts_idx_positive_white_5_prune_tlag_first_directed}
\end{subfigure}
\caption{The estimated TS GCM using \textsc{AnomalyCD} on the EasyVista dataset: a) GCMs with PDAG before edge pruning at $t=0$, and b) GCMs with DAG after edge pruning using chi-square test at $t=0$}
\label{fig:rca__easyrca_rca_ts_idx_positive_white_5}
\vspace*{-\baselineskip}
\end{figure*}

We present an ablation study in Table~\ref{tbl:rca__easyrca_anomalycd_ablation} for the \textsc{AnomalyCD} approach to demonstrate the efficacy of the additional complexity using $\bar{I}$, $\mathcal{S}_d$, $\mathcal{S}_e$ and $\Gamma_e$ methods in leveraging the PCMCI algorithm~\cite{runge2020discovering}. The \textsc{AnomalyCD} has enhanced the CD with relative gain compared to the PCMCI F$_1$ score, FPR, and SHDU by $20.5\%$, $47\%$, and $41\%$, respectively, demonstrating improved GCM accuracy. 
The relatively lower APRC ($-11.2\%$) is due to the relatively lower edge recall of \textsc{AnomalyCD}, as the spurious links in the PCMCI inflates the APRC score on the raw data. However, the gain of removing false positives outweighs the missed false negatives, as the overall performance improved in the F$_1$ and SHDU.
The correlation-based CI test can remain symmetric and fail to distinguish edge direction at $t=0$ when there is no time-lagged factor. Thus, the \textsc{AnomalyCD} is enhanced using chi-square CI test to directed the bidirected edges (see \textupdateref{line 17} in Appendix Algorithm~\ref{alg:rca__undirected_edge_pruning}). This further improves the F$_1$ is $0.381~(+26.2\%)$, FPR is $0.421~(-53.0\%)$, SHDU is $10~(-41.2\%)$, and processing time is $7.567~(-89.9\%)$.
Figure~\ref{fig:rca__easyrca_rca_ts_idx_positive_white_5} illustrates the GCM before and after directing the edges at $t=0$ using chi-square test. 

\begin{table*}[]  
\caption{Ablation study on our \textsc{AnomalyCD} approach using the EasyVista dataset. 
The \textsc{AnomalyCD}: the \textsc{PCMCI} is leveraged by sparse handling, ANAC, and edge pruning
}
\centering
\resizebox{1\linewidth}{!}{
\begin{tabular}{@{}llllll@{}}
\toprule
\textbf{Method} &  \textbf{F$_1$}$\uparrow$&  \textbf{FPR}$\downarrow$&  \textbf{APRC}$\uparrow$&   \textbf{SHDU}$\downarrow$ &   \textbf{PT (secs)}$\downarrow$
\\         
\midrule

PCMCI~\cite{runge2020discovering}                              & 0.302 (--)              & 0.895 (--)               & \textbf{0.543 (--)} & 17 (--)               & 74.667 (--)              \\
\textsc{AnomalyCD}: PCMCI + ANAC                       & 0.304 (0.7\%)           & 0.684 (-23.6\%)          & 0.499 (-8.1\%)      & 13 (-23.5\%)          & 73.366 (-1.7\%)          \\
\textsc{AnomalyCD}: PCMCI + SDH                        & 0.316 (4.6\%)           & 0.632 (-29.4\%)          & 0.460 (-15.3\%)     & 13 (-23.5\%)          & 7.004 (-90.6\%)          \\
\textsc{AnomalyCD}: PCMCI + SDH + Edge Pruning         & 0.333 (10.3\%)          & 0.632 (-29.4\%)          & 0.468 (-13.8\%)     & 13 (-23.5\%)          & 7.032 (-90.6\%)          \\
\textsc{AnomalyCD}: PCMCI + SDH + ANAC                 & 0.353 (16.9\%)          & \textbf{0.474 (-47.0\%)} & 0.477 (-12.2\%)     & \textbf{10 (-41.2\%)} & 6.571 (-91.2\%)          \\
\textsc{AnomalyCD}: PCMCI + SDH + ANAC + Edge Pruning  & \textbf{0.364 (20.5\%)} & \textbf{0.474 (-47.0\%)} & 0.482 (-11.2\%)     & \textbf{10 (-41.2\%)} & 6.596 (-91.2\%)          \\
\textsc{AnomalyCD}: PCMCI + SDLH                       & 0.316 (4.6\%)           & 0.632 (-29.4\%)          & 0.460 (-15.3\%)     & 13 (-23.5\%)          & 6.207 (-91.7\%)          \\
\textsc{AnomalyCD}: PCMCI + SDLH + Edge Pruning        & 0.333 (10.3\%)          & 0.632 (-29.4\%)          & 0.468 (-13.8\%)     & 13 (-23.5\%)          & 6.232 (-91.7\%)          \\
\textsc{AnomalyCD}: PCMCI + SDLH + ANAC                & 0.353 (16.9\%)          & \textbf{0.474 (-47.0\%)} & 0.477 (-12.2\%)     & \textbf{10 (-41.2\%)} & \textbf{6.198 (-91.7\%)} \\
\textsc{AnomalyCD}: PCMCI + SDLH + ANAC + Edge Pruning & \textbf{0.364 (20.5\%)} & \textbf{0.474 (-47.0\%)} & 0.482 (-11.2\%)     & \textbf{10 (-41.2\%)} & 6.222 (-91.7\%)         \\
\botrule
\end{tabular}
}
\label{tbl:rca__easyrca_anomalycd_ablation}
\vspace*{-\baselineskip}
\end{table*}

\begin{figure*}[h]
\centering
\includegraphics[width=0.7\linewidth]{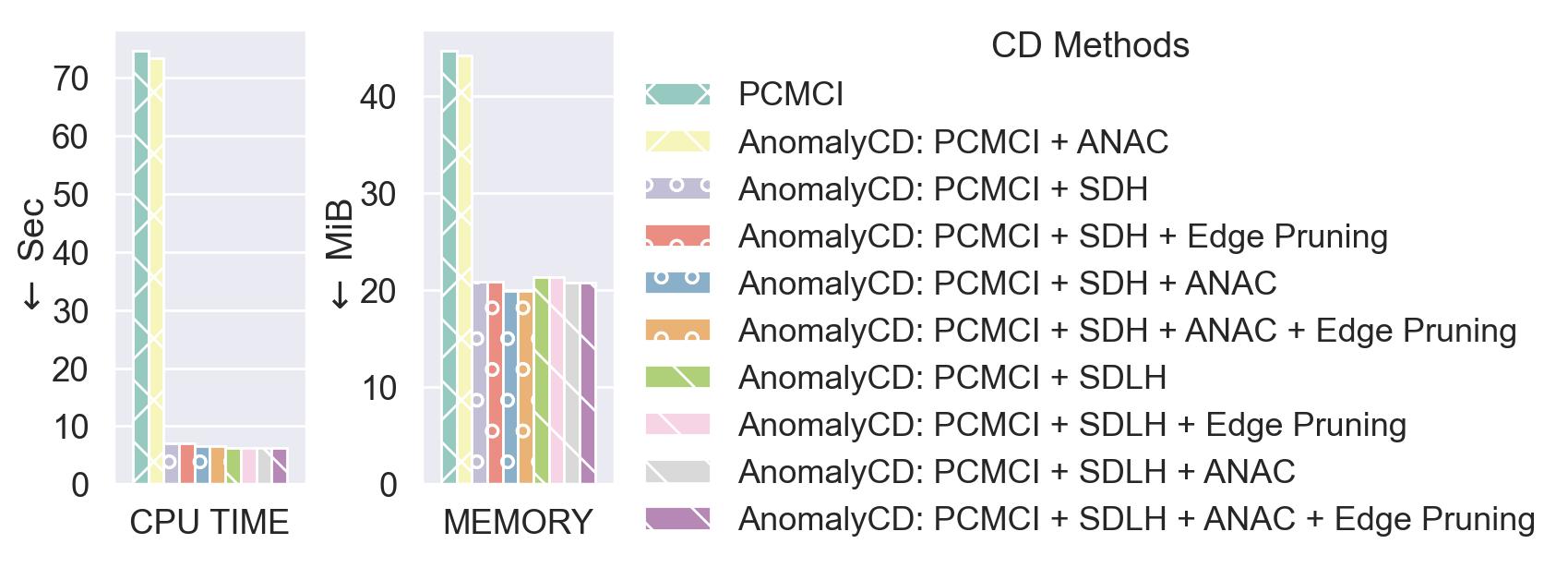}
\caption{Computational cost analysis on GCM learning on the EasyVista dataset. Our SDH $\mathcal{S}_d$ compresses the data from roughly 4,300 to 1,900 samples ($55\%\downarrow$), and the CD consumes 20~MB ($53\%\downarrow$) and executes in 7 seconds ($91\%\downarrow$). The $\mathcal{S}_e$ of the SDLH reduces the initial edges from 376 to 324 ($13.8\%\downarrow$) at $\alpha_\tau=0.10$
}
\label{fig:cc_EasyVista}
\vspace*{-\baselineskip}
\end{figure*}

Figure~\ref{fig:cc_EasyVista} depicts the computational cost of the GCM learning, and our \textsc{AnomalyCD} reduces the memory cost by half and the processing time by 10 times with $55\%$ data compression. We found the cost of the pruning $\Gamma_e$ negligible with less than $0.1\%$ increment in processing time.  

\section{Conclusion}
\label{sec:conclusion}

In this study, we have presented a novel scalable framework for discovering causal graphs from binary anomaly flag datasets. The framework incorporates various approaches to tackle the computational and accuracy challenges of inferring causality in sparse binary anomaly data.
We have systematically integrated the characteristics of anomaly flag data into causal condition testing, sparse data compression, prior link compression, and pruning adjustments to achieve state-of-the-art computation efficiency and accuracy. 
The results of our experiments have demonstrated promising efficacy in unsupervised online anomaly detection and a significant reduction in the computational overhead associated with graphical causal discovery on time series data sets. 
Our approaches will facilitate real-time diagnostic tasks across various subsystems of large systems, such as the Hadron Calorimeter and other sectors related to Industry 4.0, addressing the demand for flexible, unsupervised, and lightweight approaches that are capable of handling complex and dynamic systems with limited annotated data sets.
The tools presented in this study are deployed at CERN's Large Hadron Collider to provide anomaly detection diagnostics for the calorimeter systems.

\bmhead{Acknowledgements}

We sincerely appreciate the CMS collaboration, the HCAL operation group, the HCAL publication committee, and the CMS machine learning teams. Their technical expertise, diligent follow-up on our work, and thorough manuscript review have been invaluable.
We also thank the collaborators for building and maintaining the detector systems used in our study.
We extend our appreciation to CERN for the operations of the LHC accelerator. 
The teams at CERN have also received support from the Belgian Fonds de la Recherche Scientifique,
and Fonds voor Wetenschappelijk Onderzoek; the Brazilian Funding
Agencies (CNPq, CAPES, FAPERJ, FAPERGS, and FAPESP); SRNSF (Georgia);
the Bundesministerium f\"ur Bildung und Forschung, the Deutsche
Forschungsgemeinschaft (DFG), under Germany's Excellence Strategy --
EXC 2121 ''Quantum Universe'' -- 390833306, and under project number
400140256 - GRK2497, and Helmholtz-Gemeinschaft Deutscher
Forschungszentren, Germany; the National Research, Development and
Innovation Office (NKFIH) (Hungary) under project numbers K~128713,
K~143460, and TKP2021-NKTA-64; the Department of Atomic Energy and the
Department of Science and Technology, India; the Ministry of Science,
ICT and Future Planning, and National Research Foundation (NRF),
Republic of Korea; the Lithuanian Academy of Sciences; the Scientific
and Technical Research Council of Turkey, and Turkish Energy, Nuclear
and Mineral Research Agency; the National Academy of Sciences of
Ukraine; the US Department of Energy.

\begin{appendices}
\section{\textsc{AnomalyCD} Algorithms}
\label{sec:appendix_alg}
The appendix presents our proposed algorithms for the online AD approach and the anomaly CD methods that are discussed in Section \ref{sec:methodology}.

\begin{algorithm*}
\caption{Online time series anomaly detection}
\label{alg:rca__online_outlier_detection}
\begin{algorithmic}[1]
\scriptsize
\Procedure{OnlineTemporalAnomalyDetection}{$\mathbf{x}, p_\iota, \alpha_\theta, w_\theta, \alpha_\iota, k_\iota, \alpha_\phi, q_\phi$}
\newline
\Comment Ensemble online TS AD
\newline
\Comment $\mathbf{x}$ is a univariate TS signal data
\State $\Lambda_\theta, ~\Lambda_\iota \gets \textsc{TemporalOutlierDetection}(\mathbf{x}, p_\iota, \alpha_\theta, w_\theta, \alpha_\iota, k_\iota)$ \Comment for temporal outliers in the time domain
\State $\Lambda_\phi \gets \Phi(\mathbf{x}, \alpha_\phi, q_\phi)$ \Comment for spectral outliers in the frequency domain
\State $\Lambda \gets \Lambda_\theta \cup \Lambda_\iota \cup \Lambda_\phi$ \Comment union of binary outlier flags
\newline
\Return $\Lambda$
\EndProcedure
\end{algorithmic}
\end{algorithm*}

\begin{algorithm*}
\caption{Time-domain AD}
\label{alg:rca__online_outlier_detection_time_domain}
\begin{algorithmic}[1]
\scriptsize
\Procedure{TemporalOutlierDetection}{$\mathbf{x}, p_\iota, \alpha_\theta, w_\theta, \alpha_\iota, k_\iota$}
\newline
\Comment Temporal AD for temporal transient and gradual trend outliers in the time domain
\newline
\Comment $p_\iota$ is the window size of the convolutional filter for trend estimation
\If{$p_\iota = \text{Auto}$} \Comment for auto period estimation 
    \State $p_\iota \gets SignalPeriodEstimation(\mathbf{x})$ \Comment signal period estimation using FFT
\EndIf

\State $x_\iota, x_\zeta, x_\epsilon \gets TimeSeriesDecomposition(\mathbf{x}, p_\iota)$ \Comment decomposition into a trend, cyclic and residual
\State $\Lambda_\theta \gets \Theta(x_\epsilon, \alpha_\theta, w_\theta)$
\State $\Lambda_\iota \gets \mathcal{L}(x_\iota, \alpha_\iota, k_\iota)$
\newline
\Return $\Lambda_\theta, ~\Lambda_\iota$
\EndProcedure

\Procedure{$\Theta$}{$x_\epsilon, \alpha_\theta, w_\theta$}
\newline
\Comment Moving residual standard deviation for transient temporal AD
\newline
\Comment $\alpha_\theta$ is the detection threshold
\newline
\Comment $w_\theta$ is the sliding time-window size
\State $\lambda_\theta \gets [~]$ \Comment placeholder for outlier flag data

\For{$\mathbf{x}_\omega ~\in~GetSlicedTimeWindowData(x_\epsilon, w_\theta, step=1): $} \Comment data windowing
\State $\mu_w, ~\sigma_w \gets GetStats(\mathbf{x}_w)$ \Comment median and standard deviation using quantile $Q=[10\%, 90\%]$ 
\State $\lambda_\theta \gets Append(\lambda_\theta, |\mathbf{x} - \mu_w|/\sigma_w) $ \Comment AD score
\EndFor

\State $\Lambda_\theta \gets \lambda_\theta > \alpha_\theta$ \Comment temporal outlier flag
\newline
\Return $\Lambda_\theta$
\EndProcedure

\Procedure{$\mathcal{L}$}{$x_\iota, \alpha_\iota, k_\iota$}
\newline
\Comment Cumulative sum-based trend drift AD
\newline
\Comment $\alpha_\iota$ is the detection threshold
\newline
\Comment $k_\iota$ is a scaling constant

\State $d_\iota \gets \Delta(x_\iota): x_\iota(t) - x_\iota(t-1)$ \Comment step change of trend data points
\State $\mu _{d_\iota} \gets GetMedian(|d_\iota|)$ \Comment average step change on the trend
\State $\bar{d_\iota} \gets |d_\iota| > k_\iota \mu _{d_\iota}$ \Comment check for large step changes in the trend
\State $\lambda_\iota \gets  InitializeWithZero(\mathcal{L}(x_\iota))$ \Comment score placeholder with length of $x_\iota$, $\mathcal{L}(x_\iota)$
\For{$t_r,\bar{d_\iota}_r ~\in~GetUniformRegions(\bar{d_\iota}): $}  \Comment $\bar{d_\iota}_r \in \{0, 1\}$ is value of region $r$, and  $t_r$ is the time index of $r$
       \If{$\bar{d_\iota}_r=0$}
        \State $\lambda_\iota [t_r[j]] \gets \sum_{i=1}^{j}{d_\iota [t_r[i]]},~\text{for}~j=[1,...,N_{r}]$  \Comment drift score using cumulative sum and $N_r=\mathcal{L}({r})$
        \EndIf
\EndFor
\State $\Lambda_\iota \gets \lambda_\iota > \alpha_\iota$ \Comment trend outlier flag
\newline
\Return $\Lambda_\iota$
\EndProcedure
\end{algorithmic}
\end{algorithm*}

\begin{algorithm}
\caption{Frequency-domain AD}
\label{alg:rca__online_outlier_detection_freq_domain}
\begin{algorithmic}[1]
\scriptsize
\Procedure{$\Phi$}{$\mathbf{x}, \alpha_\phi, q_\phi$}
\newline
\Comment Spectral residual (SR) for AD
\newline
\Comment $\alpha_\phi$ is the detection threshold
\newline
\Comment $q_\phi$ is siding spectral kernel size
\State $\phi \gets SR(\mathbf{x}, q_\phi)$ \Comment outlier saliency score
\State $\lambda_\phi \gets \frac{\phi -\overline{\phi}}{\overline{\phi}}$ \Comment normalized AD score
\State $\Lambda_\phi \gets \lambda_\phi > \alpha_\phi$ \Comment spectral outlier flag
\newline
\Return $\Lambda_\phi$
\EndProcedure

\end{algorithmic}
\end{algorithm}

\begin{algorithm*}
\caption{$\mathcal{S}_d$: Data compression on sparse binary anomaly flag}
\label{alg:rca__sparse_handler}
\begin{algorithmic}[1]
\scriptsize
\Procedure{$\mathcal{S}_d$}{$\Lambda, l_m$}
\newline
\Comment Sparse (uniform no-state change) regions compression algorithm
\newline
\Comment $\Lambda \in \mathbb{R}^{N \times T}$ is a matrix of anomaly flags with $N$ sensors and $T$ time length 
\newline
\Comment $l_m$ is the maximum time length for uniform states data compression 

\State $\Lambda_c \gets [~]$ \Comment the placeholder for compressed data $\Lambda_c \in \mathbb{R}^{N \times T_c}$ 

\State $S \gets AggregateAnomalyState(\Lambda)$ \Comment returns sequence of aggregated states of from all sensors 
\For{$I, l_s \in~GetUniformRegions(S): $}   
    \Comment $I$ holds indices of the uniform region with a time length of $l_s$
    \State $\Lambda_I \gets \Lambda[I]$ \Comment $\Lambda_I$ is the selected uniform status flag region 
       \If {$l_s > l_m$}
        \State $I_d \gets GetRangeToRemove(\Lambda_I, l_m)$ 
        \Comment the tail indices of the region, excluding the first $l_m$ indices
        \State $\Lambda_I \gets CompressBinayData(\Lambda_I, I_d)$  \Comment compressing time length of $\Lambda_I$ by removing $I_d$ data indices
        \EndIf
        \State $\Lambda_c \gets Append (\Lambda_c, \Lambda_I)$
\EndFor
\newline
\Return $\Lambda_c$
\EndProcedure
\end{algorithmic}
\end{algorithm*}
 
\begin{algorithm*}
\caption{$\mathcal{S}_e$: Sparse link handling with refined prior link assumption}
\label{alg:rca__sparse_link_assum_handling}
\begin{algorithmic}[1]
\scriptsize
\Procedure{$\mathcal{S}_e$}{$\Lambda, \mathbf{E}, \tau_{\text{max}}, \alpha_\tau$}
\newline
\Comment Prior graph link preparation for GCM through removal of spurious and weak edges
\newline
\Comment $\Lambda \in \mathcal{B}^{n \times N}$ is TS binary anomaly data 
\newline
\Comment $\mathbf{E} \subseteq N \times N$ is a prior assumption of time-lagged edge links, default is all
\newline
\Comment $\tau_{\text{max}}$ is the causality search maximum time-lag
\newline
\Comment $\alpha_\tau$ is the time-lagged anomaly flag overlap strength threshold
\State $ \mathbf{E} \gets RemoveSelfTimeLagEdges(\mathbf{E})$  \Comment remove all self time-lag edges from all the nodes
\State $d\Lambda  \gets \Delta \Lambda:  \Lambda(t) -  \Lambda (t-1)$ \Comment transition of binary data
\State $S_{d\Lambda}  \gets ExtendAnomalyRegions(d\Lambda, z=\tau_{\text{max}})$   \Comment rolling sum with sliding window size $z=\tau_{\text{max}}$
\State $\Lambda_\tau \gets S_{d\Lambda} > 0$ \Comment convert to binary

\For{$\Lambda_\tau (i), \Lambda_\tau (j) ~\in~GetUniquePairSensors(\Lambda_\tau): $}     
\State $ \lambda_\tau \gets SimultaneousAnomalyCount(\Lambda_\tau (i), \Lambda_\tau (j))$ \Comment number of overlapped flags between sensor $i$ and $j$
       \If{$\lambda_\tau > 0 $} \Comment check if overlap exists
           \State $ n^i \gets AnomalyCount(\Lambda_\tau (i)) $ \Comment total number of flags for sensor $i$
           \State $ n^j \gets AnomalyCount(\Lambda_\tau (j)) $ \Comment total number of flags for sensor $j$
           
           \State $ \lambda^{ij}_\tau \gets \frac{\lambda_\tau}{n^i} $  \Comment normalized overlap score from $i$ to $j$
           \State $ \lambda^{ji}_\tau \gets \frac{\lambda_\tau}{n^j} $  \Comment normalized overlap score from $j$ to $i$
            \If {$ \lambda^{ij}_\tau < \alpha_\tau$}
                \State $ \mathbf{E} \gets RemoveEdges( \mathbf{E}, i, j)$  \Comment remove all edges from node $i$ to $j$ 
            \EndIf
             \If {$ \lambda^{ji}_\tau < \alpha_\tau$}
                \State $ \mathbf{E} \gets RemoveEdges( \mathbf{E}, j, i)$  \Comment remove all edges from node $j$ to $i$ 
            \EndIf
        \Else
            \State $ \mathbf{E} \gets RemoveEdges( \mathbf{E}, i, j)$  \Comment remove all edges from node $i$ to $j$ 
            \State $ \mathbf{E} \gets RemoveEdges( \mathbf{E}, j, i)$  \Comment remove all edges from node $j$ to $i$ 
        \EndIf
\EndFor
\newline
\Return $\mathbf{E}$
\EndProcedure
\end{algorithmic}
\end{algorithm*}

\begin{algorithm*}
\caption{$\Gamma_e$: Pruning and adjusting time-lagged causality edges}
\label{alg:rca__undirected_edge_pruning}
\begin{algorithmic}[1]
\scriptsize
\Procedure{$\Gamma_e$}{$\mathbf{E}, prune\_order=[\text{'t'},\text{'w'}]$}
\newline
\Comment Aggregates edge time lags and orients the best edges from bidirected links
\newline
\Comment $\mathbf{E} \in \mathbb{R}^{N \times 4}$ is a matrix of weighted directed time-lagged edge links $\varepsilon(\upsilon, \nu, w, t)$ 
\State $\mathbf{G}_\varepsilon \gets EdgeGroupMaxWeightedTimeLag(\mathbf{E})$  \Comment groups same-nodes edges and peak the strongest time-lag
\State $\mathbf{P} \gets [~]$ \Comment placeholder for edges to be pruned
\State $\mathbf{U} \gets [~]$ \Comment placeholder for undirected edges 
\For{$\varepsilon_s, \varepsilon_r \in~GetBidirectLinkedNodes(\mathbf{G}_\varepsilon ): $} 
     \Comment the current edge $\varepsilon_s$ and the reverse edge $\varepsilon_r$
           \State $\varepsilon_p \gets \textsc{CompareBidirectLinks}(\varepsilon_s, \varepsilon_r, m=prune\_order[0])$ \Comment get the edge to be pruned
            \If {$\varepsilon_p = \text{Null}$}
                 \State $\varepsilon_p \gets \textsc{CompareBidirectLinks}(\varepsilon_s, \varepsilon_r, m=prune\_order[1])$ \Comment get the edge to be pruned
                 \If {$\varepsilon_p = \text{Null}$}
                    \State $\mathbf{U} \gets Append(\mathbf{U}, [\varepsilon_s, \varepsilon_r])$  \Comment add the edges $\varepsilon_s$ and $\varepsilon_r$ into the undirected bucket
                      \State $\mathbf{P} \gets Append(\mathbf{P}, [\varepsilon_s, \varepsilon_r])$  \Comment add the edge $\varepsilon_s$  and $\varepsilon_r$ into prune bucket
                \Else
                    \State $\mathbf{P} \gets Append(\mathbf{P},\varepsilon_p)$  \Comment add the edge $\varepsilon_p$ into prune bucket
                \EndIf
            \Else
                \State $\mathbf{P} \gets Append(\mathbf{P},\varepsilon_p)$  \Comment add the edge $\varepsilon_p$ into prune bucket
            \EndIf

\EndFor
\State $\bar{\mathbf{E}_R} \gets PruneEdges(\mathbf{G}_\varepsilon, \mathbf{P})$  \Comment remove edges in the prune bucket
\State $\bar{\mathbf{E}_D} \gets DirectEdges(\mathbf{G}_\varepsilon, \mathbf{U})$  \Comment get DAG for the undirected edges without altering the directed edges in $\mathbf{G}_\varepsilon$ 
\State $\bar{\mathbf{E}} \gets Merge(\bar{\mathbf{E}_R}, \bar{\mathbf{E}_D})$\Comment merge pruned and directed edges 
\newline
\Return $\bar{\mathbf{E}}$
\EndProcedure

\Procedure{CompareBidirectLinks}{$\mathbf{G}_\varepsilon, \varepsilon_s, \varepsilon_r, m$}
\newline
\Comment $m=\text{'w'}$ prune based on the values of the edges' weight
\newline
\Comment $m=\text{'t'}$ prune based on the values of the edges' time lag
\State $t_s, w_s \gets GetEdgeAttributes(\varepsilon_s)$ \Comment get edge link time-lag and weight of link $s$
\State $t_r, w_r \gets GetEdgeAttributes(\varepsilon_r)$ \Comment get edge link time-lag and weight of the reverse link $r$
\If{$m = \text{'w'}$} \Comment use link weight values
    \If {$ w_s > w_r$}
        \State $\varepsilon_p \gets \varepsilon_r$ 
     \ElsIf {$ w_s < w_r$}
        \State $\varepsilon_p \gets \varepsilon_s$ 
    \Else
        \State $\varepsilon_p \gets \text{Null} $  \Comment  indicates undirected edges 
    \EndIf
        
\ElsIf{$m = \text{'t'}$} \Comment use link time-lag values (negative)
    \If {$ t_s < t_r$} 
        \State $\varepsilon_p \gets \varepsilon_r$   
    \ElsIf {$ t_s > t_r$}
        \State $\varepsilon_p \gets \varepsilon_s$   
    \Else
        \State $\varepsilon_p \gets \text{Null} $  \Comment  indicates undirected edges 
    \EndIf
\EndIf
\Return $\varepsilon_p$
\EndProcedure
\end{algorithmic}
\end{algorithm*}

\begin{algorithm*}
\caption{Temporal Bayesian network model generation}
\label{alg:rca__bn_training}
\begin{algorithmic}[1]
\scriptsize
\Procedure{BNGeneration}{$\Lambda, \mathcal{G}$}
\newline
\Comment $\Lambda \in \mathcal{B}^{n \times N}$ is multivariate TS binary anomaly data 
\newline
\Comment $\mathcal{G}$ is a graph network with $\mathcal{G}(\upsilon,\nu, \varepsilon(w, t))$
\State $\hat{\mathcal{G}} \gets UnrollTemporalGraph(\mathcal{G})$ \Comment update node names with edge time-lag value, Eq.~\ref{eq:unroll_graph}
\newline
\Comment The loop below is for TS data unrolling, adopted from TPC~\cite{biswas2022statistical}
\State $\hat{\Lambda} \gets [~]$   \Comment empty place holder for the unrolled TS data
\For{$\varepsilon(\upsilon_t, \nu) \in~GetLinkedNodes(\mathcal{G}): $}   
        \State $t \gets GetEdgeTimeLag(\upsilon_t)$ \Comment get source node time-lag 
       \If{$t < 0$} \Comment checks the time delayed causality
        \State $x_\upsilon \gets \Lambda[\upsilon]$ \Comment get TS data of the $\upsilon$ sensor, $x_\upsilon \in \mathbb{R}^{1 \times T}$
        \State $\hat{x}_\upsilon \gets GetTimeDelayedData(x_\upsilon, t)$  \Comment shifts backward the TS $x_\upsilon$ by $t$ time steps
        \State $\hat{\Lambda} \gets Append(\hat{x}_\upsilon)$ 
        \EndIf
\EndFor

\State $\mathcal{M}_b \gets FitBayesianNetwork(\hat{\Lambda}, \hat{\mathcal{G}})$  \Comment build BN model
\newline
\Return $\mathcal{M}_b$
\EndProcedure

\end{algorithmic}
\end{algorithm*}

\section{Further Results on Causal Graph Discovery}
\label{sec:rca__causal_results}
\subsection{Causal Graph Discovery on the HCAL Dataset}
\label{sec:rca__causal_hcal_data_multi_lm}

\begin{figure*}[!htbp]
\centering
\begin{subfigure}[]{1\linewidth}
\centering
\includegraphics[width=1\linewidth]{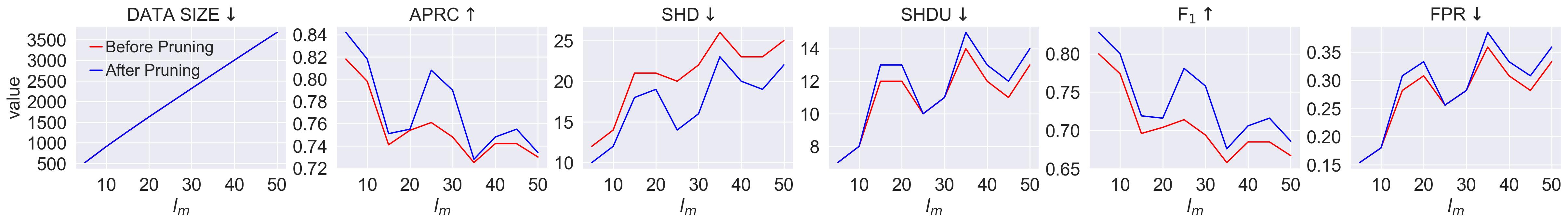}
\caption{}
\label{fig:CMS_HCAL_HEP07_rca_issparsedataTrue_issparselinkFalse_refGDomian_score_compare_df}
\end{subfigure}
\begin{subfigure}[]{1\linewidth}
\centering
\includegraphics[width=1\linewidth]{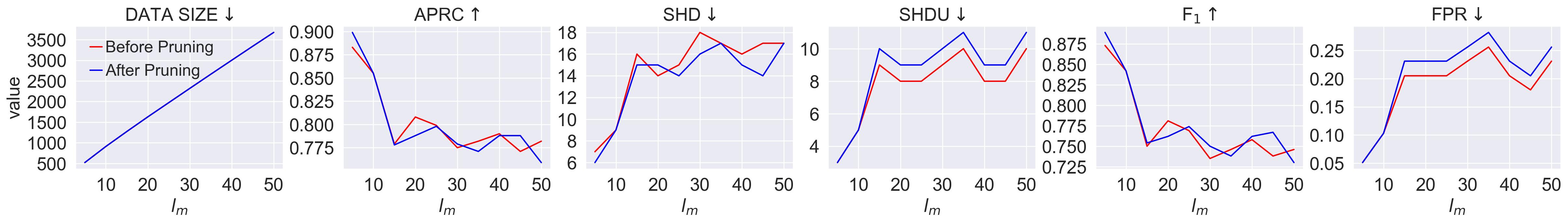}
\caption{}
\label{fig:CMS_HCAL_HEP07_rca_issparsedataTrue_issparselinkTrue_refGDomian_score_compare_df}
\end{subfigure}
\caption{\textupdate{Causal graph learning comparison on $\tau_\text{max}=5$ and different sparsity length $l_m={5, 10, \dots, 50}$: a) with sparse data handling, and b) with sparse data and link handling. The accuracy scores are measured with respect to the domain GCM of the HCAL given in Fig.~\ref{fig:rca__hcal_rm_ground_truth}.}}
\label{fig:CMS_HCAL_HEP07_rca_issparsedataTrue_issparselinkBoth_refGDomian_score_compare_df}
\vspace*{-\baselineskip}
\end{figure*}

The appendix presents a performance comparison of CD using different sparsity lengths $l_m$ on the HCAL dataset. Figure~\ref{fig:CMS_HCAL_HEP07_rca_issparsedataTrue_issparselinkBoth_refGDomian_score_compare_df} illustrates the data compression and CD performance trend with $\mathcal{S}_d + \mathcal{S}_e + \Gamma_e$ for $l_m={5,10,\dots,50}$. The results demonstrate a promising and consistent gain with $\mathcal{S}_e$ + $\Gamma_e$. However, we have observed varying CD accuracy across the $l_m$'s, which is attributed to the dependency of the partial correlation CI test on data size. The value of $l_m$ affects data compression, resulting in varying data sizes and relative frequencies of $0$s and $1$s in the binary data. The lower $l_m=\tau_\text{max}=5$ achieves the best score, as spurious links increase for longer $l_m$ due to higher sparsity attribution with uniform binary sequences longer than the $\tau_\text{max}$. 

\end{appendices}


\bibliography{bibi.bib}

\newpage
\onecolumn
\section*{The CMS-HCAL Collaboration}
\author{M.W.~Asres\cmsorcid{0000-0003-2985-108X}}$^{1}$, 
\author{C.W.~Omlin\cmsorcid{0000-0003-0299-171X}}$^{1}$, 
\author{V.~Chekhovsky}$^{2,a}$, 
\author{A.~Chinaryan}$^{2}$, 
\author{A.~Gevorgyan\cmsorcid{0000-0003-2751-9489}}$^{2}$, 
\author{Ya.~Halkin}$^{2}$, 
\author{A.~Kunts}$^{2}$, 
\author{A.~Litomin}$^{2,a}$, 
\author{A.~Petrosyan}$^{2}$, 
\author{A.~Tumasyan\cmsorcid{0009-0000-0684-6742}}$^{2}$
% ...


\vspace{4ex}
\noindent $^{1}$University~of~Agder, Grimstad, Norway\cmsAuthorMark{*}\\
$^{2}$Yerevan~Physics~Institute, Yerevan, Armenia\\
\begin{flushleft}
% ...
$^{*}$No longer in CMS HCAL Collaboration\\
$\dag$:~Deceased\\
$^{a}$Also at an institute or an international laboratory covered by a cooperation agreement with CERN\\
% ...
\end{flushleft}

\maketitle

\end{document}